\documentclass[letterpaper]{article}
\usepackage{style}
\usepackage[hyphens]{url}
\usepackage{graphicx}
\usepackage{natbib}
\usepackage{caption}
\usepackage{algorithm}
\usepackage{algorithmic}

\usepackage{times}
\usepackage{helvet}
\usepackage{courier}
\usepackage[hyphens]{url}
\usepackage{graphicx}
\usepackage{booktabs}
\usepackage{multirow}
\usepackage{amsmath}
\usepackage{xspace}
\usepackage[x11names]{xcolor}
\usepackage{colortbl}
\usepackage{arydshln}
\definecolor{lightblue}{RGB}{218, 232, 252}
\definecolor{lightgray}{RGB}{220,220,220}
\definecolor{lightpurple}{RGB}{230,230,255}
\definecolor{mygray}{gray}{0.92}
\definecolor{groveblue}{HTML}{DDE7F3} 

\usepackage[most]{tcolorbox}
\definecolor{cPattern}{HTML}{7E3FA1}
\definecolor{cEpisode}{HTML}{C2185B}
\definecolor{cMoment}{HTML}{2E7D32}

\newtcbox{\pill}[1]{on line, nobeforeafter, boxrule=0pt, boxsep=0pt,
  left=2.5pt, right=2.5pt, top=0.8pt, bottom=0.8pt, arc=2.2pt,
  colback=#1, colframe=#1, fontupper=\bfseries\scriptsize\color{white}}

\newcommand{\pattern}{\pill{cPattern}{PATTERN}}
\newcommand{\episode}{\pill{cEpisode}{EPISODE}}
\newcommand{\moment}{\pill{cMoment}{MOMENT}}

\newcommand{\method}{GROVE\xspace}            
\newcommand{\registry}{perceptual trace} 

\usepackage{newfloat}
\usepackage{listings}
\DeclareCaptionStyle{ruled}{labelfont=normalfont,labelsep=colon,strut=off}
\floatstyle{ruled}
\newfloat{listing}{tb}{lst}{}
\floatname{listing}{Listing}

\tcbuselibrary{listings, breakable, skins}
\lstdefinestyle{promptstyle}{
  basicstyle=\ttfamily\scriptsize, breaklines=true, breakatwhitespace=false,
  breakautoindent=false, breakindent=0pt, columns=fullflexible, keepspaces=true,
  showstringspaces=false, numbers=none, xleftmargin=4pt, framexleftmargin=4pt,
  aboveskip=1pt, belowskip=6pt, upquote=false,
  frame=single, framerule=0.4pt, rulecolor=\color{black!45},
  backgroundcolor=\color{black!3}
}

\usepackage{booktabs}

\title{GROVE: Growing and Reasoning over Temporally Stratified Memory from Streaming Video Experience}
\author{
    Sitong Gong\textsuperscript{\rm 1,2}, Caixin Kang\textsuperscript{\rm 2,3}\equalcontrib, Tianyu Yan\textsuperscript{\rm 1,2}\equalcontrib, Guo Chen\textsuperscript{\rm 4}, Bo Zheng\textsuperscript{\rm 2}, Kaipeng Zhang\textsuperscript{\rm 2}, Yunzhi Zhuge\textsuperscript{\rm 1}, Xiang Ruan\textsuperscript{\rm 1}, Huchuan Lu\textsuperscript{\rm 1}, Yifei Huang\textsuperscript{\rm 2,3}\corresponding
}
\affiliations{

    \textsuperscript{\rm 1}Dalian University of Technology 
    \textsuperscript{\rm 2}Alaya Lab 
    \textsuperscript{\rm 3}The University of Tokyo 
    \textsuperscript{\rm 4}NVIDIA \\
    stgong@mail.dlut.edu.cn, hyf015@gmail.com
}

\begin{document}

\maketitle

\begin{abstract}
A wearable assistant should both answer questions about its visual history and recognize when that history is useful to the present situation. Existing video-memory systems primarily support question-conditioned recall, whereas proactive assistants typically use separate memory and control mechanisms. We introduce GROVE, a training-free framework that supports both behaviors with one memory grown causally from a continuous video stream. GROVE retains fine-grained perceptual evidence and incrementally consolidates it into time-stamped moments, coherent episodes, and recurring cross-day patterns. Each stratum is paired with a scale-native retrieval skill for locating an observation, replaying an activity, or traversing long-range regularities. Reactive QA and proactive assistance share this memory and access interface, differing in whether retrieval is initiated by a user query or the current situation. Across multiple benchmarks including the challenging MM-lifelong and EgoServe, GROVE achieves the best results among the compared methods. Controlled ablations show that the temporal strata and their access skills are complementary, with patterns providing the largest benefit when evidence spans multiple days. 
Code will be available at \url{https://github.com/SitongGong/GROVE}.
\end{abstract}

\section{Introduction}
A wearable assistant observes the world from its user's perspective for hours, days, or even weeks. To remain helpful over this growing history, it must support two complementary forms of assistance. When a user asks, ``Where did I leave my keys yesterday?'', the assistant must \emph{pull} relevant evidence from the past. When the user is about to repeat a mistake, it should instead \emph{push} timely assistance without being asked. Reactive question answering and proactive assistance therefore depend on the same underlying capability: turning an open-ended video stream into a memory that remains faithful, searchable, and actionable over time.

\begin{figure}
    \centering
    \includegraphics[width=0.95\linewidth]{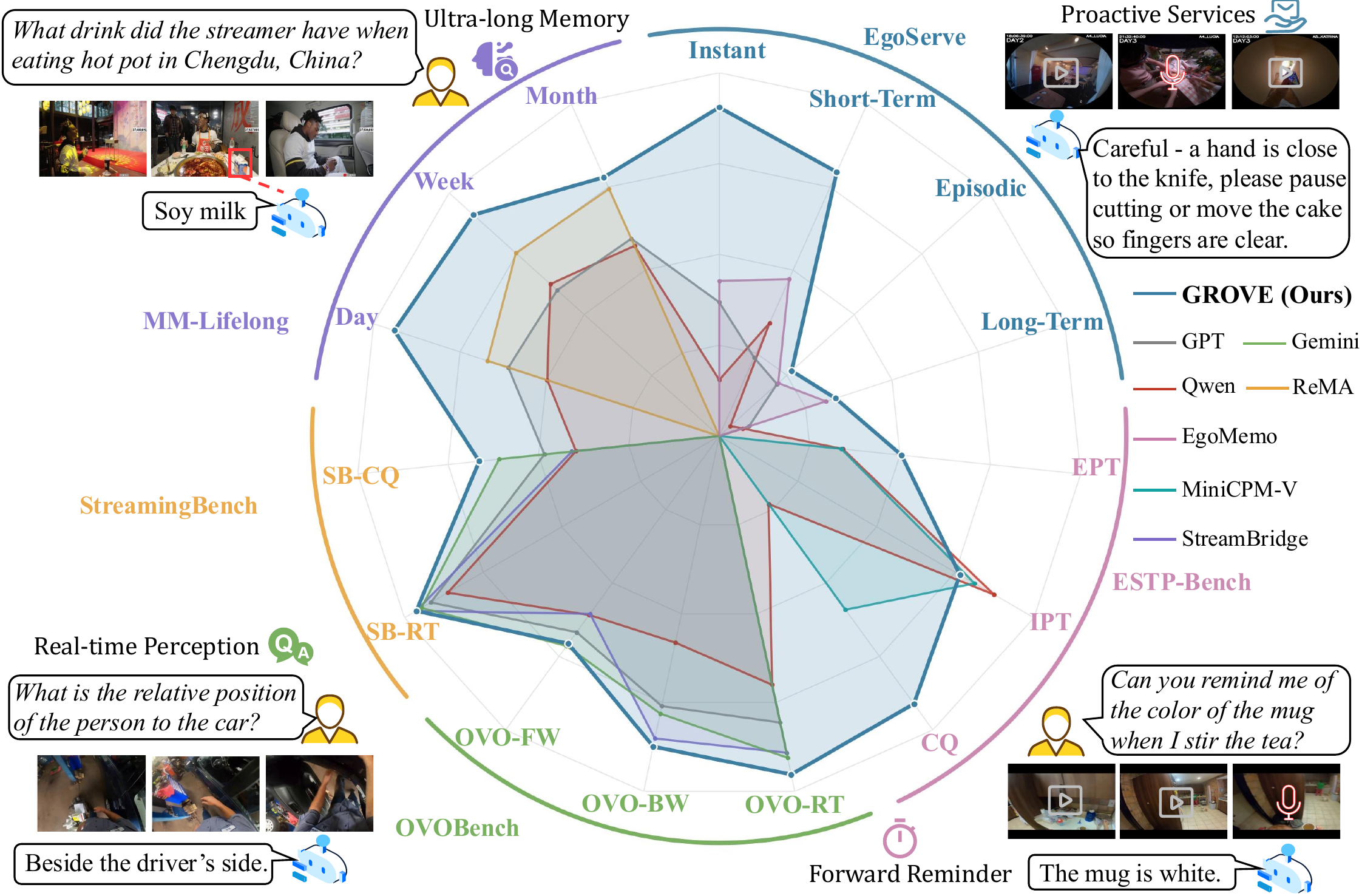}
    \vspace{-2mm}
    \caption{Performance of GROVE across five video understanding benchmarks. With a temporally structured memory, GROVE answers questions spanning ultra-long horizons, stays grounded in real-time perception, and decides when to deliver proactive services or forward reminders.  
    }
    \label{fig:teasor}
    \vspace{-5mm}
\end{figure}

Recent video-memory systems already address many difficulties of long and streaming video. They construct online event memories, organize experience hierarchically, and use iterative or agentic retrieval to locate sparse evidence \cite{xie2026streamrag,long2025seeing,li2026visual,chen2026memdreamer,li2026bridging,liang2026oasis}. These advances change how efficiently the past is stored and searched, but most retain the same \emph{query-first} contract: an explicit request determines what should be retrieved. Proactive assistants instead use the current situation to decide whether intervention is warranted \cite{zhang2025eyes,sitong2026vinci2}, but their memory and control mechanisms are designed primarily for service triggering rather than as a common interface for open-ended recall. The missing capability is therefore not merely another memory level or a stronger search procedure. It is a single streaming memory that can be accessed from either of two starting signals: a question seeking past evidence or the unfolding situation indicating that past experience may now be useful.

Our key idea is to make temporal scale the interface between what memory stores and how an agent reads it. Evidence required by a video assistant varies qualitatively with scale: an object count or state change may last seconds, an activity may unfold over minutes, and a routine may emerge only after related activities recur across days. Treating these observations as homogeneous records forces one general search operation to recover fundamentally different kinds of evidence. In GROVE, each scale instead represents a distinct kind of information and exposes an access operation suited to that information. The hierarchy is therefore not used only to compress the video; it determines how the stored experience can be reasoned over. Long-range regularities are explicitly constructed because they are absent from any individual frame or event, and all memory updates are causal so that the memory remains usable while the stream is still arriving.

We realize this idea in \textbf{GROVE} (\textbf{G}rowing and \textbf{R}easoning \textbf{O}ver Streaming \textbf{V}ideo \textbf{E}xperience), a training-free framework that grows a temporally stratified memory online. Each arriving video window is processed through two complementary perception channels: a narrative caption and a structured perceptual registry containing entities, attributes, explicit counts, actions, and on-screen text. These observations are consolidated into time-stamped moments, coherent episodes, and cross-episode patterns capturing recurring behavior. GROVE exposes four scale-native retrieval skills (\emph{Perception Lookup}, \emph{Moment Recall}, \emph{Episode Replay}, and \emph{Pattern Traversal}) that enter the memory at increasing temporal scope. A multi-round agent composes these skills and accumulates non-redundant evidence.


The same memory and skill library support two control policies. For \textit{reactive} question answering, the user query initiates a decide--retrieve--answer loop in which the agent selects the type and scale of evidence to retrieve. For \textit{proactive} assistance, either a forward-looking request or the current perceptual state initiates retrieval of related moments, episodes, and patterns before the agent decides whether and when to act. The two policies read the same constructed past through the same scale-native interface.


We evaluate GROVE on MM-Lifelong~\cite{chen2026towards}, OVO-Bench~\cite{niu2025ovo}, StreamingBench~\cite{lin2026streamingbench}, ESTP-Bench~\cite{zhang2025eyes}, and EgoServe~\cite{sitong2026vinci2}, covering lifelong question answering, online video understanding, forward assistance, and query-free proactive services. GROVE achieves the best overall results among the compared methods on all three benchmarks. Controlled ablations show that both the temporal organization of memory andits scale-aware retrieval interface contribute to these gains. The full design also improves retrieval efficiency, requiring fewer reasoning rounds and lower latency than a flat caption index.
 
Our contributions are threefold:
\begin{itemize}
\item We formulate long-video memory as a shared substrate for reactive question answering and proactive assistance.
\item We introduce GROVE, a \textbf{causal, temporally stratified memory} paired with scale-native retrieval skills.
\item Across 5 benchmarks, we demonstrate the effectiveness of this shared memory and retrieval interface, with ablations validating both its memory structure and retrieval design.
\end{itemize}

\section{Related Works}

\noindent\textbf{Streaming Video Understanding.}
Offline long-video models extend the context window or compress an entire
clip before answering \cite{chen2025longvila,song2024moviechat,azad2025hierarq},
an assumption that fails when the video is still arriving. Online video LLMs
instead process each window as it comes and learn when to respond
\cite{chen2024videollm,wang2024videollm,zhang2024flash}; recent systems
refine this with disentangled perception--decision loops
\cite{qian2025dispider}, redundancy-aware token pruning
\cite{yao2025timechat}, persistent event memory
\cite{zeng2026streamforest,wang2026streambridge}, and visual instruction
feedback \cite{fu2025vispeak}. Progress in this setting is measured by
OVO-Bench \cite{niu2025ovo}, which probes real-time perception, backward
tracing and forward active responding, and StreamingBench
\cite{lin2026streamingbench}, which adds contextual understanding; both are
part of our evaluation. Such models maintain a compact state for the
immediate present, whereas we retain a structured memory that stays
queryable over days.
 
\noindent\textbf{Memory-based Video Agents.}
A complementary, training-free line of work equips a frozen MLLM with an
external memory and retrieves from it on demand. Retrieval-augmented
pipelines index captions or clips of the whole video
\cite{ren2026videorag,luo2026video}, while agentic systems add tool use and
multi-round search over that index
\cite{zhang2026deep,liu2025videomind,yang2026longvt}. Closest to us are
memories with explicit structure: VAM commits deduplicated moments into an
age-layered store \cite{li2026visual}, M3-Agent maintains entity-centric
episodic and semantic memory from audio-visual streams
\cite{long2025seeing}, StreamRAG segments events online for query-adaptive
retrieval \cite{xie2026streamrag}, and MemDreamer
\cite{chen2026memdreamer} and MAGIC-Video \cite{li2026bridging} pair
hierarchical or multimodal memory graphs with an agentic tool bank; ReMA
shows that dynamic memory management becomes essential at the lifelong
horizon \cite{chen2026towards}. Our stratification echoes the classic
episodic--semantic distinction in human memory \cite{tulving1972}, but
differs in three respects: the memory is grown causally online, it is
stratified by temporal scale with one retrieval skill per stratum, and it
serves proactive assistance in addition to question answering.
 
\noindent\textbf{Proactive Assistants.}
Proactivity was first studied in text-only dialogue, where an agent must
decide not only what to say but when to speak unprompted
\cite{liu2025proactive}. In egocentric video it becomes a streaming decision
problem: ProAssist generates task-guidance dialogue from the stream
\cite{zhang2025proactive}, Eyes Wide Open times synchronized proactive
answers and introduces ESTP-Bench \cite{zhang2025eyes}, and Vinci2 defines
the service-triggering protocol over EgoServe \cite{sitong2026vinci2}.
These systems tie the trigger decision to current perception; none is built
on a memory that also serves reactive reasoning, and none exposes the
cross-day structure that long-term services require.

\begin{figure*}
    \centering
    \includegraphics[width=0.93\linewidth]{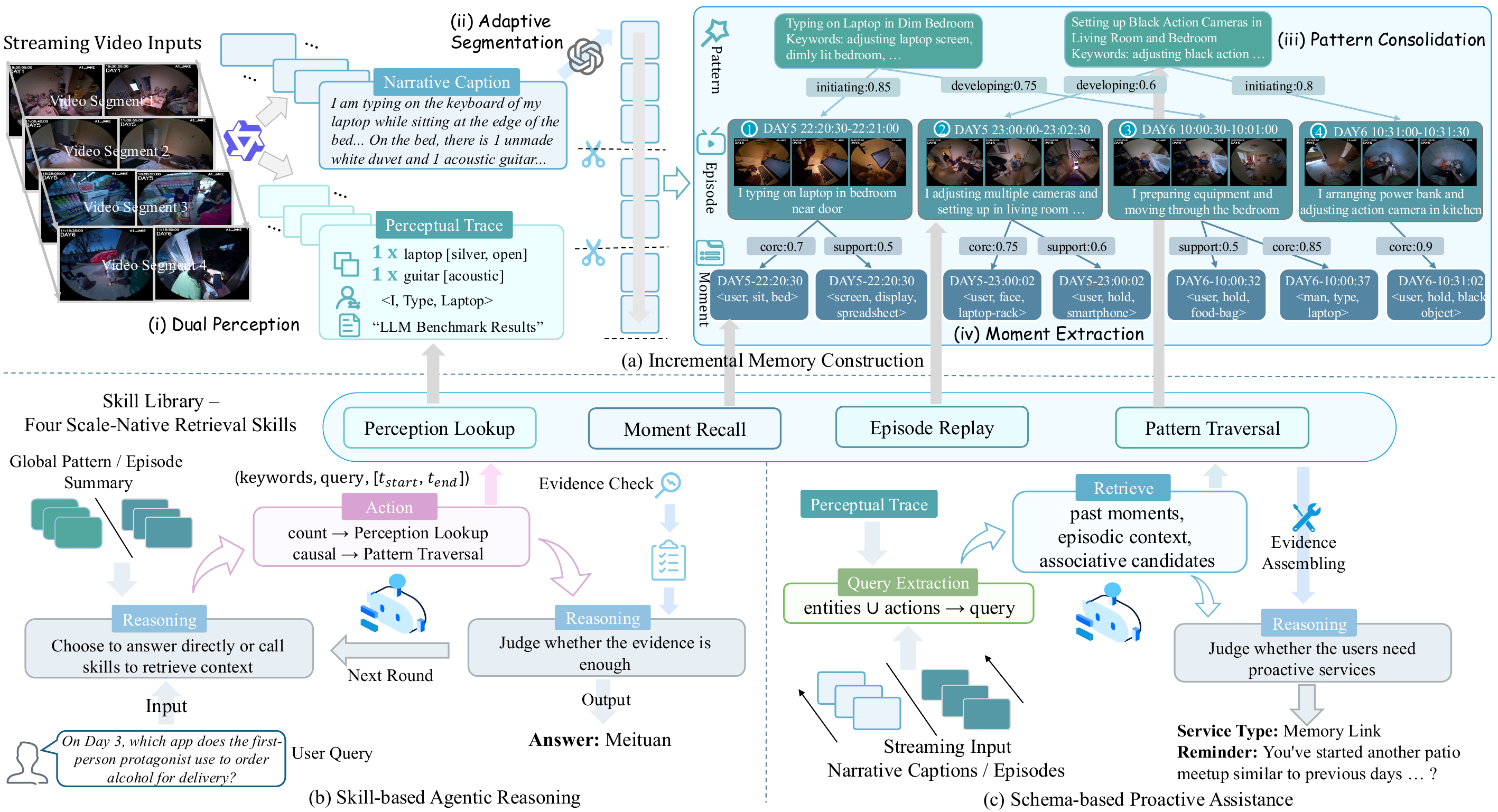}
    \vspace{-1mm}
    \caption{Overview of GROVE. (a) Memory construction proceeds in four incremental steps: dual perception, adaptive segmentation, pattern consolidation and moment extraction. (b) Four retrieval skills expose the strata to an agent that iteratively judges whether the retrieved evidence suffices to answer a user query. (c) Given no user query, the agent instead invokes the same skills to recall contextual clues from the memory, and decides on that evidence whether a proactive assistance is warranted.}
    \vspace{-3mm}
    \label{fig:pipeline}
\end{figure*}

\section{Methodology}

\noindent\textbf{Problem setting.}
We represent a streaming video as consecutive windows $V=\{w_1,w_2,\ldots,w_t\}$. At time $t$, the system can access only the observed prefix and its memory $\mathcal{M}_{\leq t}$. GROVE supports two tasks. In \emph{reactive question answering}, a query $q$ requests evidence from the observed history. In \emph{proactive assistance}, the system receives either no request or a forward-looking request $q^+$ and must decide whether the current situation warrants a response:
\begin{equation}
y=\pi\!\left(q,\ \mathcal{M}_{\le t}\right),\qquad
s_t=\pi\!\left(q^{+},\ \mathcal{M}_{\le t}\right),
\label{eq:tasks}
\end{equation}
where $\pi$ is the reasoning agent, $q^{+}$ may be absent, and $s_t$ is
either empty or a response anchored at a second $\tau\!\le\!t$.

\noindent\textbf{Overview.}
\method{} decouples the system into streaming memory construction and streaming memory access by agentic reasoning (Fig.~\ref{fig:pipeline}). As video arrives, a perception
model and a lightweight consolidation model grow a textual memory stratified by temporal scale:
\begin{equation}
\mathcal{M}_{\le t}=\{\mathcal{R},\ \mathcal{F},\ \mathcal{E},\ \mathcal{P}\},
\label{eq:memory}
\end{equation}
where $\mathcal{R}$ is the perceptual trace and $\mathcal{F}$, $\mathcal{E}$, and $\mathcal{P}$ contain \moment{}s, \episode{}s, and \pattern{}s. Each stratum exposes one scale-native retrieval skill, forming a skill library $\mathcal{A}=\{a_1,\dots,a_4\}$. Both tasks in
Eq.~\ref{eq:tasks} are then solved by the same agent over the same memory through the same skills: $\pi$ invokes skills in $\mathcal{A}$ over multiple rounds until the retrieved evidence suffices, and never revisits raw frames.
 
\subsection{Incremental Memory Construction}
\label{sec:construction}
We build the stratified memory in four incremental steps. Each arriving
window first undergoes dual perception, producing dense narrative captions
and a \registry{} entry. An adaptive segmenter groups the streaming captions
into \episode{}s. When an \episode{} closes, it is atomized into
time-stamped \moment{}s and consolidated with related, temporally
non-contiguous \episode{}s into \pattern{}s. Every operation updates the
previous state $\mathcal{X}_{t^-}$ using only observations available by time
$t$; no retrospective pass over future video is required. The resulting
memory can therefore be queried while the stream is still arriving.
The current window is immediately available through $\mathcal{R}$; \moment{}s,
\episode{}s, and \pattern{}s are updated only after the open \episode{} closes.


\noindent\textbf{Dual perception.}
Each input window $w_{t}$ is parsed by a VLM $\Phi$ into two
complementary views $(c_t,\ r_t)$,
where $c_t$ contains timestamped captions of sampled frames and a short window-level summary. The \registry{} entry $r_t$ stores three types of timestamped evidence: object counts and attributes, subject--action--object events, and transcribed on-screen text. The \registry{} supplements captions with explicit entities, counts, and text
that narrative descriptions tend to blur. Both views are retained, $\mathcal{R}_t=\mathcal{R}_{t^-}\cup\{(c_t,r_t)\}$, yielding a time-ordered record of the observed stream.

\noindent\textbf{Adaptive Episode Segmentation.}
The captions $c_t$ are streamed into an LLM segmenter
$\mathrm{Seg}$ that judges whether the ongoing activity continues:
\begin{equation}
b_t=\mathrm{Seg}\big(c_t,\ \tilde{e}_{t^-}\big)\in\{\textsc{extend},\ \textsc{cut}\},
\label{eq:segment}
\end{equation}
where $\tilde{e}_{t^-}$ denotes the \episode{} currently being accumulated.
The segmenter matches its anchors against the arriving captions and remains
conservative, so that ambiguous evidence keeps the activity open. On
\textsc{extend} the window is absorbed into the open unit and the activity
keeps growing, until either a semantic change is detected or a maximum-length
cap is reached. On \textsc{cut} the unit is closed and appended,
$\mathcal{E}_t=\mathcal{E}_{t^-}\cup\{e_t\}$, where the closed \episode{} $e$ stores its start and end times together with two LLM-written texts: a brief summary of the activity, and a longer description that retains its fine-grained spatio-temporal details; a cut is also forced at a large recording gap or a day change. Episodes are therefore variable-length units aligned with activity boundaries rather than fixed slices, which keeps each unit semantically coherent and gives retrieval a meaningful span to land on.

\noindent\textbf{Moment Extraction.}
Whenever an \episode{} closes, a lightweight consolidation model $\Psi$
atomizes its dense perception into time-stamped \moment{}s $\mathcal{F}_t$, formulated as: 
\begin{equation}
\{f_i\}_{i=1}^{n}=\Psi(e_t),\qquad
\mathcal{F}_t=\mathcal{F}_{t^-}\cup\{f_i\}_{i=1}^{n},
\label{eq:moment}
\end{equation}
where $\mathcal{F}_t$ contains all \moment{}s produced by time $t$. Each $f_i$ pairs a timestamp within $e_t$ with a minimal statement of an utterance, state change, or action, plus search keywords. The same pass creates a group edge from $e_t$ to its \moment{}s and assigns each \moment{} a role (\emph{core}, \emph{context}, or \emph{detail}) and a salience weight. The edge lets retrieval recover the surrounding \episode{} in one hop and favor core facts over peripheral detail.

 
\noindent\textbf{Pattern Consolidation.}
Finally, an LLM matches the closed \episode{} against existing \pattern{}s using its title, summary, and keywords:
\begin{equation}
p^{*}=\mathrm{Match}\big(e_t,\ \mathcal{P}_{t^-}\big),
\label{eq:match}
\end{equation}
where $p^*$ is empty if no candidate accepts $e_t$. The stratum is then updated as
\begin{equation}
\mathcal{P}_t=
\begin{cases}
\big(\mathcal{P}_{t^-}\!\setminus\!\{p^{*}\}\big)\cup\{\mathrm{Upd}(p^{*},e_t)\}, & p^{*}\ \text{found},\\[2pt]
\mathcal{P}_{t^-}\cup\{\mathrm{New}(e_t)\}, & \text{otherwise},
\end{cases}
\label{eq:pattern}
\end{equation}
Here, $\mathrm{Upd}$ absorbs $e_t$ and re-synthesizes the pattern title and summary from its episodic history, whereas $\mathrm{New}$ opens a candidate pattern. A \pattern{} is stored as a group edge over temporally non-contiguous \episode{}s; after multiple \episode{}s have been assigned, it explicitly records recurrence, frequency, and typical time. This is the only stratum that represents regularities absent from any single \episode{}.

\subsection{Scale-Native Retrieval Skills}
\label{sec:primitives}
 


GROVE exposes four \emph{scale-native} retrieval skills, one per stratum (Table~\ref{tab:skills}). Because the strata store different evidence at different granularities, each skill enters the memory at the requested scale instead of ranking all record types in one flat index. A skill may provide several calls, but all calls share a source stratum and return granularity. They accept keywords or a semantic query and, when available, a time range. Together, the four skills define the action space $\mathcal{A}$. A benchmark may expose only the subset required by its task.

\begin{table}[t]
\centering
\caption{The retrieval skill library of GROVE.}
\vspace{-3mm}
\label{tab:skills}
\setlength{\tabcolsep}{4pt}
\begin{tabular}{@{}llp{3.3cm}@{}}
\toprule
Skill & Reads & Best for \\
\midrule
Perception Lookup & $\mathcal{R}$ & exact counts, attributes, on-screen text at a moment \\
Moment Recall & $\mathcal{F}$ & what/when facts by keyword and time range \\
Episode Replay & $\mathcal{E}$ & locating and reading an activity segment \\
Pattern Traversal & $\mathcal{P}{\to}\mathcal{E}{\to}\mathcal{F}$ & broad/causal/cross-period queries \\
\bottomrule
\end{tabular}
\vspace{-3mm}
\end{table}

\noindent\textbf{Perception Lookup.}
This skill reads $\mathcal{R}$ at a requested time and returns nearby captions or structured entity counts, attributes, events, and on-screen text. It serves fine-grained questions and grounds real-time queries in the present scene.
 
 
\noindent\textbf{Moment Recall.}
This skill searches $\mathcal{F}$ by keyword and time span, returning matching time-stamped moments or the ordered sequence within an interval. It addresses what-happened-when questions without replaying an entire \episode{}.

\noindent\textbf{Episode Replay.}
This skill searches \episode{} summaries or reads recent units in $\mathcal{E}$, then returns the selected activity's detailed description and span $[t_0,t_1]$. The span can constrain subsequent Perception Lookup or Moment Recall calls, providing a coarse-to-fine path.
 
 

\noindent\textbf{Pattern Traversal.}
This skill follows group edges along $\mathcal{P}{\to}\mathcal{E}{\to}\mathcal{F}$, with each stage constraining the candidates at the next. It is useful when a broad or cross-period query does not specify a time range. At each stage, reciprocal rank fusion combines BM25 and dense semantic rankings so that both named entities and paraphrased concepts remain reachable. Traversal may also begin at $\mathcal{E}$ or $\mathcal{F}$ when $\mathcal{P}$ is unnecessary.

 
 
\subsection{Retrieval over the Memory}
\label{sec:heads}
Both modes in Eq.~\ref{eq:tasks} access memory through $\mathcal{A}$. Every call respects the causal cutoff and exposes only memory constructed by the current time $t$. The policies differ in how retrieval is initiated and
controlled.

\noindent\textbf{Skill-based Agentic Reasoning.}
Given a reactive query (Fig.~\ref{fig:pipeline}b),
$\pi$ runs a bounded loop of at most $T_{\max}$ rounds. At round
$k$, the agent either answers from the accumulated evidence
$\mathcal{C}_{k-1}$ or invokes a skill $a_k\in\mathcal{A}$. The returned
evidence is deduplicated before being appended:
\begin{equation}
o_k=a_k(\mathcal{M}_{\leq t}),\qquad
\mathcal{C}_k=\mathcal{C}_{k-1}\cup(o_k\setminus\mathcal{C}_{k-1}).
\label{eq:dedup}
\end{equation}
Query cues guide the initial scale: counts and literal text favor Perception
Lookup, specific events favor Moment Recall, activities favor Episode Replay,
and broad cross-period questions favor Pattern Traversal. The agent may move
coarse-to-fine across rounds and stops when the evidence is sufficient. For
real-time questions, the current perceptual trace is included before retrieval.

\noindent\textbf{Schema-based Proactive Assistance.}
Given a forward-looking query or no query at all
(Fig.~\ref{fig:pipeline}c), there is no past-oriented user query to route.
Instead, $\pi$ converts the entities and actions in the current
\registry{} into a retrieval query. A fixed schema then fills three evidence
slots: preceding \moment{}s, the surrounding \episode{}, and associated
\pattern{}s that represent routines or earlier commitments. The assembled
evidence is used to decide whether a service is needed and, if so, what to
provide. When every \episode{} must be screened, scale-matched experts evaluate
service families in parallel: immediate safety and guidance at the moment
scale, reminders at the episode scale, and longer-term coaching at the pattern
scale. A fired service is aligned to the perception grid for second-level
timing.

\begin{table*}[t]
\centering
\caption{Proactive service performance on EgoServe with best results per column in \textbf{bold}.}
\vspace{-3mm}
\label{tab:egoserve}
\scalebox{0.9}{
\begin{tabular}{l|cc|ccc|cc|ccc|c}
\toprule
\multirow{2}{*}{Model} & \multicolumn{2}{c|}{Instant} & \multicolumn{3}{c|}{Short-term} & \multicolumn{2}{c|}{Episodic} & \multicolumn{3}{c|}{Long-term} & \multirow{2}{*}{Overall} \\
 & SA & TU & NSG & ER & RR & MR & TR & HC & ML & RO & \\
\midrule
Qwen3-VL-Plus~\cite{bai2025qwen3} & 5.2 & 1.5 & 8.6 & \underline{10.4} & 3.4 & 0.0 & 1.8 & 4.4 & 0.0 & 0.0 & 3.5 \\
GPT-5-mini~\cite{singh2025openai} & 12.5 & 3.6 & 9.5 & 1.0 & \underline{5.2} & 0.0 & \textbf{9.4} & \underline{5.7} & 0.0 & 0.0 & 4.7 \\
EgoMemo~\cite{sitong2026vinci2} & \underline{11.4} & \underline{7.5} & \textbf{24.7} & 1.7 & 4.7 & \underline{3.8} & 5.7 & 3.7 & \underline{4.9} & \textbf{11.8} & \underline{8.0} \\
\midrule
\rowcolor{groveblue}
GROVE (Ours) & \textbf{19.7} & \textbf{20.2} & \underline{21.8} & \textbf{23.8} & \textbf{6.8} & \textbf{5.9} & \underline{5.9} & \textbf{8.6} & \textbf{6.5} & 7.1 & \textbf{12.6} \\
\bottomrule
\end{tabular}}
\vspace{-3mm}
\end{table*}
\begin{table}[t]
\centering
\small
\caption{Performance comparison on the MM-Lifelong.
}
\vspace{-3mm}
\label{tab:mmlifelong}
\scalebox{0.9}
{
\begin{tabular}{lccc}
\toprule
Method & 
Month
& 
Week
& 
Day
\\
\midrule
\textcolor{gray}{Human} & \textcolor{gray}{80.4} & \textcolor{gray}{95.6} & \textcolor{gray}{99.2} \\
\midrule
GPT-5~\cite{singh2025openai} & 14.87 & 15.00 & 15.25 \\
Qwen3-VL-235B-A22B~\cite{bai2025qwen3} & 14.33 & 15.63 & 12.44 \\
Video-XL-2-8B~\cite{qin2025video} & 9.07 & 12.00 & 9.00 \\
Eagle-2.5-8B~\cite{chen2026eagle} & 6.10 & 7.00 & 8.25 \\
Nemotron-v2-12B~\cite{deshmukh2025nvidia} & 9.63 & 11.00 & 7.25 \\
VideoMind-7B~\cite{liu2025videomind} & 8.35 & 11.75 & 7.50 \\
LongVT-7B~\cite{yang2026longvt} & 7.54 & 9.75 & 7.00 \\
DeepVideoDiscovery~\cite{zhang2026deep} & 10.57 & 9.02 & 10.25 \\
ReMA~\cite{chen2026towards} & \underline{18.62} & \underline{18.82} & \underline{16.75} \\
\midrule
\rowcolor{groveblue}
GROVE (ours) & \textbf{19.98} & \textbf{22.75} & \textbf{23.50} \\
\bottomrule
\end{tabular}
}
\vspace{-3mm}
\end{table}

\section{Experiments}

\subsection{Experimental Setup}

\textbf{Benchmarks and Evaluation.} We evlauate our method on 5 representative benchmarks. 
\textbf{EgoServe}~\cite{sitong2026vinci2} contains over 3,000 proactive service instances across ten service sub-types, measured with Macro-F1 and an LLM-judged quality score.
\textbf{MM-Lifelong}~\cite{chen2026towards} comprises 181.1 hours of video
across day-, week-, and month-scale splits, evaluating lifelong
question answering under the official GPT-5 judge protocol. \textbf{OVO-Bench}~\cite{niu2025ovo} evaluates online
video comprehension across backward tracing, real-time perception, and
forward active responding. \textbf{StreamingBench}~\cite{lin2026streamingbench}
assesses real-time and contextual video understanding under a
streaming constraint. 
\textbf{ESTP-Bench}~\cite{zhang2025eyes} evaluates ego-proactive video understanding through temporally grounded questions
timed at opportune moments, scored with the official
\textit{validScoreF1} metric averaged per task type.

\noindent\textbf{Implementation details.}
Perception and caption generation use Qwen3.5-35B-A3B
~\cite{team2026qwen3}, while memory consolidation uses GPT-4.1-mini. The reasoning backbone is GPT-5.2 for MM-Lifelong and GPT-5-mini for the other benchmarks. Local video processing runs on NVIDIA H200 GPUs. For MM-Lifelong and EgoServe, we sample one frame every 5s and cap episodes at 10min; for the remaining benchmarks, we sample one frame every 2s and cap episodes at 2min. Dataset-specific configurations and all construction and inference prompts are provided in the supplementary material.

\begin{table*}[t]
\centering
\caption{Online video understanding on OVO-Bench and
StreamingBench. We report the overall score of
each track (OVO real-time/backward/forward and their overall;
StreamingBench real-time, contextual, and their average). The results of Online MLLMs are reported from \cite{azad2026streamready}. 
}
\label{tab:online_qa}
\vspace{-3mm}
\resizebox{0.85\textwidth}{!}{%
\begin{tabular}{l|cccc|ccc}
\toprule
\multirow{2}{*}{Model} & \multicolumn{4}{c|}{OVO-Bench} & \multicolumn{3}{c}{StreamingBench} \\
 & Real-Time & Backward & Forward & Overall & Real-Time & Contextual & Avg \\
\midrule
\textcolor{gray}{Human Agents} & \textcolor{gray}{93.20} & \textcolor{gray}{92.33} & \textcolor{gray}{92.90} & \textcolor{gray}{92.81} & \textcolor{gray}{91.46} & \textcolor{gray}{93.55} & \textcolor{gray}{92.51} \\
\midrule
\multicolumn{8}{c}{\textit{Proprietary MLLMs}} \\
\midrule
GPT-4o~\cite{hurst2024gpt} & 64.5 & 60.8 & 53.4 & 59.5 & 73.3 & 38.7 & 56.0 \\
Claude 3.5 Sonnet~\cite{anthropic2024claude35sonnet} & $-$ & $-$ & $-$ & $-$ & 72.4 & 37.7 & 55.1 \\
\midrule
\multicolumn{8}{c}{\textit{Open-Source Offline MLLMs}} \\
\midrule
InternVL2~\cite{chen2024far} & 60.7 & 44.0 & 45.4 & 50.0 & 63.7 & 32.4 & 48.1 \\
LLaVA-OneVision~\cite{li2024llava} & 62.8 & 45.0 & 50.9 & 52.9 & 71.1 & 32.7 & 51.9 \\
Qwen2-VL~\cite{wang2024qwen2} & 56.0 & 46.7 & 48.7 & 52.7 & 69.0 & 31.7 & 50.4 \\
LLaVA-NeXT-Video~\cite{li2024llava} & 63.3 & 41.7 & 54.2 & 53.1 & 69.8 & 34.3 & 52.1 \\
VITA-1.5~\cite{fu2026vita} & 63.5 & 41.5 & 53.5 & 52.8 & 52.3 & 27.4 & 39.9 \\
HierarQ~\cite{azad2025hierarq} & 67.3 & 48.3 & 48.3 & 54.6 & 69.7 & 35.1 & 52.4 \\
\midrule
\multicolumn{8}{c}{\textit{Open-Source Online MLLMs / Agents}} \\
\midrule
VideoLLM-online~\cite{chen2024videollm} & 20.8 & 17.7 & $-$ & 19.3 & 36.0 & 26.6 & 31.3 \\
Flash-VStream~\cite{zhang2024flash} & 29.9 & 25.4 & 44.2 & 33.2 & 23.2 & 24.1 & 23.7 \\
Dispider~\cite{qian2025dispider} & 54.6 & 36.1 & 34.7 & 41.8 & 67.6 & 33.6 & 50.6 \\
StreamForest~\cite{zeng2026streamforest} & 61.2 & 52.0 & 53.5 & 55.6 & \textbf{77.3} & $-$ & $-$ \\
StreamBridge~\cite{wang2026streambridge} & \underline{71.3} & \underline{68.1} & 48.4 & \underline{62.6} & \underline{77.0} & 32.6 & 54.8 \\
ViSpeak~\cite{fu2025vispeak} & 66.3 & 57.5 & \underline{54.3} & 59.4 & 70.4 & \underline{43.9} & \underline{57.2} \\
TimeChat-Online~\cite{yao2025timechat} & 58.6 & 42.0 & 36.4 & 45.7 & 75.4 & 35.3 & 55.4 \\
StreamAgent~\cite{yang2025streamagent} & 61.3 & 41.7 & 45.4 & 49.5 & 74.3 & 34.6 & 54.5 \\
\midrule
\rowcolor{groveblue}
GROVE (Ours) & \textbf{76.2} & \textbf{69.9} & \textbf{56.5} & \textbf{67.5} & \underline{77.0} & \textbf{53.1} & \textbf{65.1} \\
\bottomrule
\end{tabular}}
\vspace{-3mm}
\end{table*}

\begin{table}[t]
\centering
\caption{Experimental results on the ESTP-Bench. EyeWO$^{*}$ is the benchmark's own model (gray). Best non-gray per column in \textbf{bold}.}
\vspace{-3mm}
\label{tab:estp_overall}
\scalebox{0.83}{
\begin{tabular}{l|cccc}
\toprule
Model & EPT & IPT & CQ & Overall \\
\midrule
\textcolor{gray}{EyeWO$^{*}$~\cite{zhang2025eyes}} & \textcolor{gray}{23.6} & \textcolor{gray}{52.5} & \textcolor{gray}{43.6} & \textcolor{gray}{34.7}  \\
\midrule
LLaVA-OneVision~\cite{li2024llava} & 13.6 & 31.8 & 8.2 & 18.0 \\
Qwen2-VL~\cite{wang2024qwen2} & 15.4 & \textbf{39.3} & 10.4 & 21.3 \\
MiniCPM-V~\cite{yao2024minicpm} & 15.2 & \underline{36.5} & \underline{26.6} & \underline{22.9} \\
LLaVA-NeXT-Video~\cite{li2024llava} & \underline{16.5} & 34.6 & 13.8 & 21.3 \\
InternVL-V2~\cite{chen2024far}  & 5.6 & 6.7 & 5.7 & 5.9 \\
LIVE~\cite{chen2024videollm}  & 9.5 & 25.6 & 18.9 & 15.5 \\
MMDuet~\cite{wang2024videollm}  & 9.1 & 34.2 & 20.3 & 17.8\\
\midrule
\rowcolor{groveblue}
GROVE (Ours) & \textbf{22.7} & 34.4 & \textbf{41.0} & \textbf{28.6} \\
\bottomrule
\end{tabular}
}
\vspace{-3mm}
\end{table}

\subsection{Main Results}

\noindent\textbf{Egoserve. }
GROVE achieves the best overall macro-F1 on EgoServe (12.6), exceeding EgoMemo by 4.6 points and ranking first on seven of ten service subtypes (Table~\ref{tab:egoserve}). Its largest gains occur in safety (19.7 vs. 12.5), tool use (20.2 vs. 7.5), and error recovery (23.8 vs. 1.7), which depend on fine-grained current and recent evidence. GROVE also improves habit
coaching (8.6 vs. 3.7) and memory-linked reminders (6.5 vs. 4.9), where cross-episode context is useful.

\noindent\textbf{MM-Lifelong.}
On MM-Lifelong (Table~\ref{tab:mmlifelong}), GROVE gets the best
score on all three horizons: 23.50 (day), 22.75 (week), and 19.98
(month), improving over the strongest
agentic baseline ReMA by 6.75, 3.93, and 1.36 points, respectively. The gain is largest on the day split, where the perceptual trace and time-stamped moments preserve details such as counts, attributes, text, and action timing. The smaller but consistent gains at longer horizons indicate that this fine-grained evidence remains accessible after consolidation.

\noindent\textbf{OVO-Bench and StreamingBench.}
On OVO-Bench and StreamingBench (Table~\ref{tab:online_qa}), GROVE obtains the best OVO-Bench overall score (67.5) and StreamingBench average (65.1) among the compared models. It leads OVO real-time perception (76.2), while obtaining 69.9 on backward tracing and 56.5 on forward active responding. On StreamingBench, it achieves 53.1 on contextual understanding, 9.2 points above the strongest listed baseline, while remaining competitive on real-time understanding (77.0 vs. 77.3). These results show that retrieval over the observed history improves temporally contextual questions without sacrificing grounding in the current window.
 
\noindent\textbf{ESTP-Bench.}
On ESTP-Bench (Table~\ref{tab:estp_overall}), GROVE achieves the best contextual-question score among the non-benchmark-specific comparison methods (41.0 vs. 26.6) and the best overall score (28.6 vs. 22.9). It also leads the explicit proactive track (22.7), but trails Qwen2-VL on implicit proactive tasks (34.4 vs.\ 39.3). The large contextual gain is consistent with GROVE's design: these questions benefit most from connecting the present to earlier events.

\begin{table}[t]
\centering
\caption{Ablation of memory structure. Each masked memory layer is replaced by an equal volume of raw captions, so that all rows receive the same information and differ only in structural organization.}
\label{tab:abl_structure}
\vspace{-3mm}
\scalebox{0.9}{
\begin{tabular}{l ccc}
\toprule
\multirow{2}{*}{Config} & \multicolumn{2}{c}{MMLifelong} & \multirow{2}{*}{EgoServe} \\
\cmidrule(lr){2-3}
   & day & week & \\
\midrule
\rowcolor{groveblue}
\textbf{GROVE} & \underline{18.75} & \underline{19.75}  & \textbf{12.62} \\
\midrule
w/o Perceptual Trace & 15.50 & \textbf{20.50}   & 12.27 \\
w/o Moment & 17.09 & 18.25  & 11.77 \\
w/o Episode & 14.57 & 17.25  & \underline{12.47} \\
w/o Pattern & \textbf{19.90} & 13.00  & 11.49 \\
Flat (w/o Hier.) & 14.75 & 15.50  & 11.53 \\
\bottomrule
\end{tabular}
}
\vspace{-3mm}
\end{table}

\begin{table}[t]
\centering
\caption{Ablation of retrieval skills. On EgoServe, the schema injects evidence blocks
instead of exposing tools, so removing a skill amounts to removing its stratum for perceptual trace and pattern layer. }
\vspace{-3mm}
\label{tab:abl_primitive}
\scalebox{0.9}{
\begin{tabular}{l cc c}
\toprule
\multirow{2}{*}{Config} & \multicolumn{2}{c}{MMLifelong} & \multirow{2}{*}{EgoServe} \\
\cmidrule(lr){2-3}
 & day & week &  \\
 \midrule
\rowcolor{groveblue}
\textbf{GROVE} & \underline{18.75} & \underline{19.75} & \textbf{12.62} \\
\midrule
w/o Perception Lookup & 16.25 & \textbf{20.25} & \underline{12.27} \\
w/o Moment Recall & \textbf{19.50} & 15.75 & 11.05 \\
w/o Episode Replay & 18.34 & 19.60 & 11.36 \\
w/o Pattern Traversal & 18.50 & 15.58 & 11.49 \\
No retrieval & 7.25 & 12.75 & 10.39 \\
\bottomrule
\end{tabular}
}
\vspace{-3mm}
\end{table}

\begin{table}[t]
\centering
\caption{Ablation of retrieval details on different benchmarks.}
\label{tab:abl_details}
\vspace{-3mm}
\scalebox{0.9}{
\begin{tabular}{l cc c}
\toprule
\multirow{2}{*}{Config} & \multicolumn{2}{c}{MMLifelong} & \multirow{2}{*}{EgoServe} \\
\cmidrule(lr){2-3}
 & day & week &  \\
 \midrule
\rowcolor{groveblue}
\textbf{GROVE} & \textbf{18.75} & \textbf{19.75} & \textbf{12.62} \\
\midrule
w/o time boundary & \underline{15.08} & 13.50 & - \\
Question Retri. & 12.56 & \underline{18.00}  & - \\
LLM-generated query & - & - & \underline{12.17} \\
\bottomrule
\end{tabular}
}
\vspace{-3mm}
\end{table}

\begin{table}[t]
\centering
\caption{Ablation of the memory strata on MM-Lifelong (day). The inference budget is computed over the same 20 questions as in Fig.~\ref{fig:efficiency}}
\label{tab:abl_ladder_mm}
\vspace{-3mm}
\scalebox{0.9}{
\begin{tabular}{lcccc}
\toprule
 Level & day & tok/q & rounds & lat.(s) \\
\midrule
caption flat & 12.75  & 15.1K & 6.25 & 59.3 \\
+\,Episode & 13.75  & 18.2K & 6.40 & 59.1 \\
+\,Moment/fact & 13.57  & 19.1K & 5.05 & 48.1 \\
+\,Pattern & 15.50 & 29.7K & 5.60 & 50.9 \\
\rowcolor{groveblue}
Full  & \textbf{18.75}  & 26.5K & \textbf{4.95} & \textbf{47.6} \\
\bottomrule
\end{tabular}
}
\end{table}

\subsection{Ablations}
We conduct ablations on MM-Lifelong day/week and EgoServe. MM-Lifelong tests reactive reasoning over long-horizon memory, whereas EgoServe tests proactive assistance. To control evaluation cost, MM-Lifelong ablations use GPT-5-mini rather than the GPT-5.2 backbone in Table~\ref{tab:mmlifelong}; all variants within an ablation share the same backbone and retrieval budget.

\noindent\textbf{Memory structure.}
Table~\ref{tab:abl_structure} replaces each stratum in turn. Removing the
\episode{} layer causes the largest day-scale drop
(18.75$\to$14.57), while flattening the hierarchy reduces performance to
14.75/15.50 on day/week despite preserving an equal volume of captions. The
effect is scale-dependent: removing \pattern{} slightly improves the day split
(18.75$\to$19.90), where cross-day recurrence is unnecessary, but sharply
reduces the week split (19.75$\to$13.00). On EgoServe, removing \moment{} causes
the largest single-stratum decline (12.62$\to$11.77).
 
\noindent\textbf{Retrieval skills.}
Table~\ref{tab:abl_primitive} deletes one skill at a time while its content
remains reachable. Disabling retrieval entirely collapses accuracy
(18.75$\to$7.25 on day), and removing \emph{Moment Recall} or \emph{Pattern
Traversal} sharply lowers the week split (15.75 / 15.58) and EgoServe
(11.05); on the day split the remaining skills largely compensate, as the
episode skeleton already answers coarse questions. Read against the
structural ablation, this separates \emph{content value} from
\emph{entry-point value}: a stratum helps only when both its content and a
native way to reach it are present.
 
\noindent\textbf{Retrieval mechanism.}
Table~\ref{tab:abl_details} varies how the skills query the memory.
Removing the temporal-range constraint degrades localization
(18.75$\to$15.08 on day, 19.75$\to$13.50 on week), and building the
proactive query from the current perceptual trace outperforms an
LLM-written query on EgoServe (12.62 vs.\ 12.17): grounding retrieval in
observed entities and actions is more reliable than free-form query
generation.

\noindent\textbf{Structure and efficiency.}
Starting from a flat caption index, we add one stratum back at a time with
the retrieval budget fixed, and measure accuracy together with the retrieved
tokens, realized rounds, and latency per question
(Table~\ref{tab:abl_ladder_mm}). Accuracy increases monotonically as the
hierarchy is restored (12.75$\to$18.75), while the agent needs \emph{fewer}
rounds (6.25$\to$4.95) and less time (59.3$\to$47.6\,s) to answer. Structure
therefore buys accuracy and efficiency at once: a memory organized by
temporal scale lets the agent reach the right evidence directly instead of
searching its way toward it.

\subsection{Inference-Time Analysis}
We analyse costs on both sides of the system: the retrieval budget in the agentic reasoning stage and the inference time in the memory construction stage. 
 
\begin{figure}
    \centering
    \includegraphics[width=1.0\linewidth]{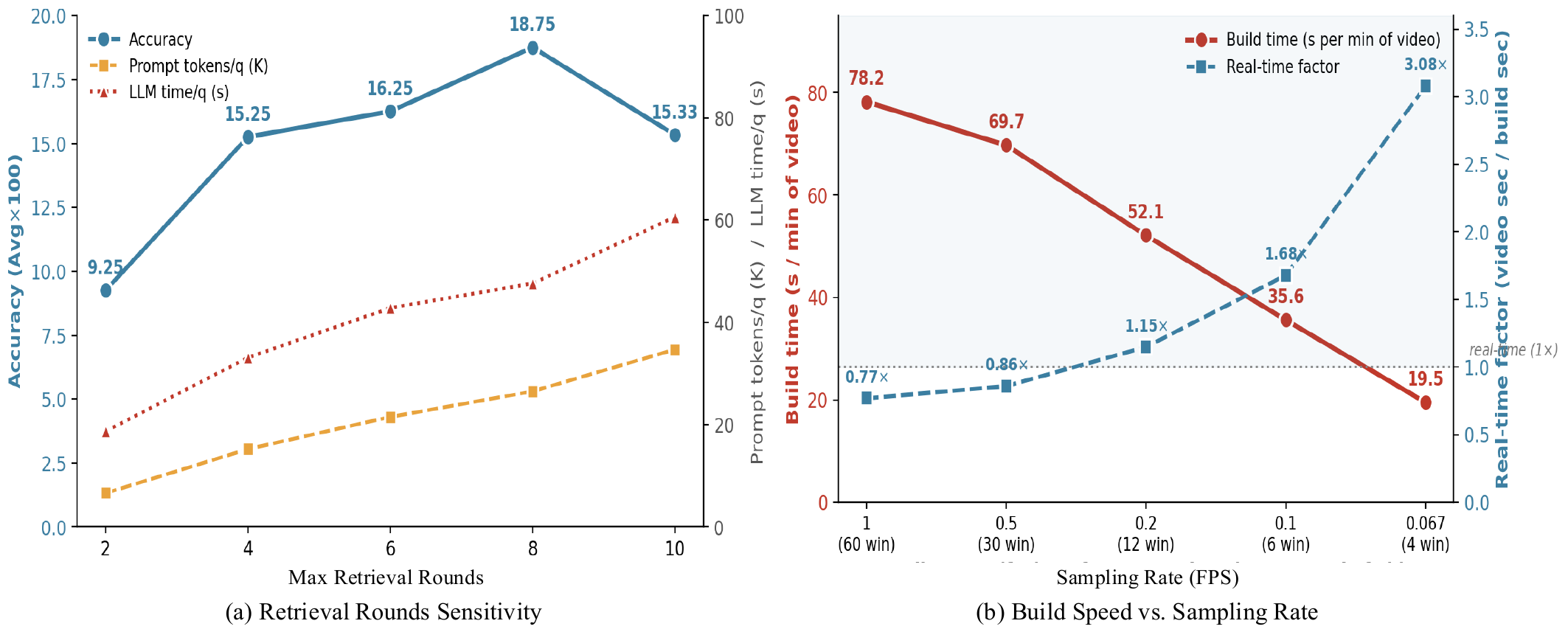}
    \caption{Inference and construction cost. (a) Accuracy, total prompt tokens, and LLM time per question as the maximum retrieval budget varies on MM-Lifelong day. (b) Build time per minute of video and the resulting real-time factor across sampling rates.}
    \label{fig:efficiency}
\end{figure}

\noindent\textbf{Retrieval budget.}
To explore how the retrieval budget affects accuracy and cost, we vary the maximum number of retrieval rounds with the memory fixed. Accuracy is measured on the full day split, while the prompt tokens and LLM time per question are profiled on a batch of 20 questions. 
As illustrated in Fig.~\ref{fig:efficiency} (a), accuracy peaks at eight rounds (18.75) and
declines beyond it (15.33 at ten), whereas tokens and latency grow almost
linearly with the budget. Additional retrieval thus stops paying off once
the evidence is sufficient-further calls mostly add low-information
context-so we set the budget to eight rounds throughout, where the agent
self-terminates after 4.95 rounds on average.
 
\noindent\textbf{Memory construction.}
To accelerate construction, we run captioning, episode segmentation, and moment/pattern consolidation as three overlapping asynchronous stages. Fig.~\ref{fig:efficiency} (b) reports the build time per minute of video across sampling rates and window counts: it drops from 78.2 to 19.5 seconds, exceeding real time from one frame every five seconds onward and reaching 3.08× at the sparsest setting. Construction can thus keep pace with a live stream, trading temporal resolution for throughput without altering the memory structure.
 
\subsection{Qualitative Analysis}
\begin{figure}
    \centering
    \includegraphics[width=0.9\linewidth]{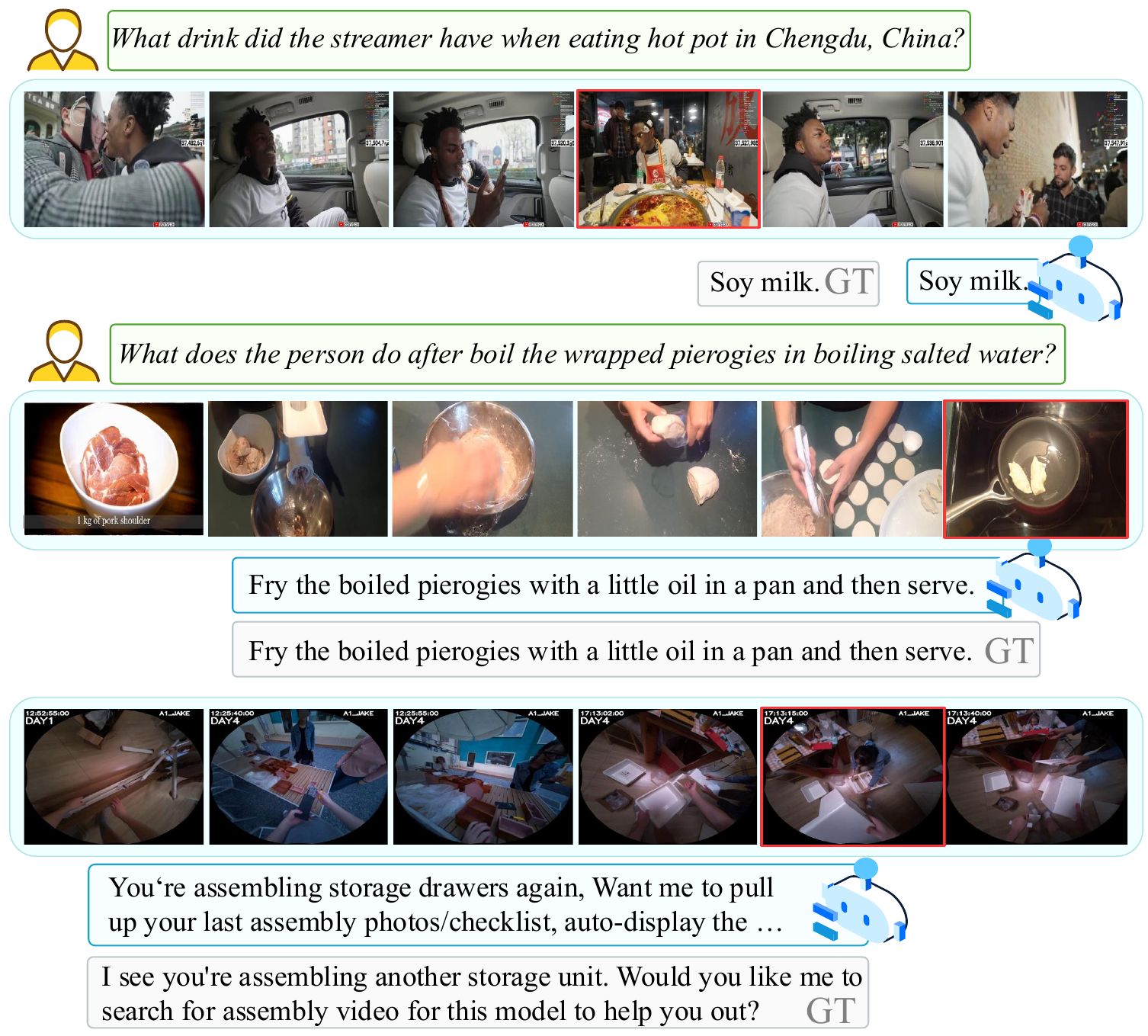}
    \vspace{-2mm}
    \caption{Visualization of GROVE's predictions. The first two cases answer user queries about a fine-grained detail and a procedural step from a long-past video, while the third fires a proactive service without any query.}
    \vspace{-4mm}
    \label{fig:visualization}
\end{figure}

Fig.~\ref{fig:visualization} shows three cases. In the first, GROVE locates the hot-pot episode among hours of unrelated footage and reads the drink from its perceptual trace, matching the ground truth. In the second, a procedural query, it replays the cooking episode and follows its ordered moments to recover the next step. In the third, with no query at all, it traverses the pattern linking the Day-4 assembly to the Day-1 session and offers the earlier record. The cases show one memory read at the right scale for both reactive and proactive use.

\section{Conclusion}
We presented GROVE, a training-free framework that organizes streaming video
memory and its access around temporal scale. GROVE causally transforms
incoming observations into a perceptual trace, time-stamped moments, coherent
episodes, and recurring patterns, with a retrieval skill specialized for each
stratum. Reactive QA and proactive assistance share this memory and skill
library. Results across five
benchmarks, together with controlled ablations, show that pairing temporal
structure with scale-native access improves both long-horizon recall and
situation-aware assistance. GROVE therefore provides a practical memory
substrate for assistants that must continuously observe, remember, and act. 

\section{Limitations}
GROVE inherits errors from its frozen perception model, and evidence missed during ingestion cannot be recovered through later consolidation. Higher-level memory is updated only when an episode closes, so current-window queries rely on the perceptual trace. Memory construction and multi-round retrieval also become more costly as the observed history grows. Finally, because the main benchmarks use different reasoning backbones, component-level conclusions rely on the fixed-backbone ablations rather than absolute cross-benchmark results.

\section*{A. Methods}
\subsection*{A.1 Details on Memory Construction}
This section complements the construction pipeline described in the main paper
with the implementation-level rules, the hyperparameters of every stage, and
the dataset-specific instantiations that we use to build all memories reported
in the experiments. 
Table~\ref{tab:supp_hparams} lists all hyperparameters for both memory construction and retrieval stages, while Algorithm~\ref{alg:construction} summarizes the online procedure.
 
\noindent\textbf{Window Perception.}
A stream is consumed as fixed-length windows. Within a window, we sample frames at a constant interval and issue a \emph{single} VLM call that returns one detailed caption per sampled frame, each with its own time span, plus one window-level description; the same call emits the \registry{} entry of that window. The sampling interval is increased from $2$s to $5$s on MM-Lifelong and EgoServe, whose recordings run for tens of hours: the number of VLM calls grows linearly with duration, so a denser interval would make these two settings account for most of the construction cost. A window incorporating captions and \registry{} is also the smallest unit the memory ever exposes---its perception becomes retrievable as soon as the VLM call returns, without waiting for the surrounding \episode{} to close.
 
\noindent\textbf{Deterministic Rules around the Segmenter.}
The LLM segmenter receives the frame-level captions of the arriving window
together with those of the three most recent windows of the open unit and makes a binary judgment on the window as a whole.
It compares stable anchors across the two: whether the high-level activity
continues, whether the location has changed, and whether the key objects and people involved are still present. A boundary is declared only when such a change
holds for the entire window, and the open unit is then closed
\emph{before} it, which becomes the first window of the next unit.
Three deterministic rules bound the judgment: a cut is forced without consulting
the LLM at recording gaps above $60$s, at day changes, and at the episode length
cap; the LLM is not queried until the open unit holds a minimum number of frames,
so a boundary is never declared from a single caption; and an unparsable output
defaults to \textsc{extend}, keeping the activity open rather than fragmenting it.
 
\noindent\textbf{Adaptive Moment Density.}
The number of \moment{}s requested from a closed \episode{} is a target rather than a fixed rate: it is set to $\lceil \text{duration}/\Delta \rceil$, with $\Delta=5$s for units up to $5$min, $10$s for units between $5$ and $15$min, and $20$s beyond that, and the extraction 
may return up to $1.5\times$ that target, with a floor of $5$ and a hard ceiling of $200$ per \episode{}.
The timestamps are not uniformly spaced---they fall wherever something actually happens---so the target only controls how densely coverage is requested: it keeps a ten-minute unit from collapsing into a handful of sentences while preventing a pathological unit from flooding the stratum. The role and weight of each \moment{} are assigned within the same extraction stage, so no additional pass over the memory is needed to build the group edge.
 
\noindent\textbf{Bounded Cost of Pattern Consolidation.}
Three measures keep consolidation from growing with the number of \pattern{}s already stored. Candidates are examined in parallel batches of $10$ and the first affirmative judgment absorbs the \episode{}, so a match short-circuits the remaining candidates. When a \pattern{} is re-synthesized, only its most recent $5$ \episode{}s contribute full summaries, and older ones are compressed into a single short subject each, which bounds the prompt regardless of how long the \pattern{} has been running. 
Roles and weights of the grouped \episode{}s are assigned default values on
insertion and re-estimated by a dedicated LLM pass only once every $3$ updates
of that \pattern{}, together with the coherence score.
 
\noindent\textbf{Dataset-specific Instantiations.}
Construction is otherwise identical across benchmarks, with three exceptions that we state explicitly for fairness.
(i) Caption and extraction prompts are designed specifically to scenario---generic third-person QA, first-person tool manipulation, and first-person daily life---so that the perception stage attends to the evidence each domain requires.
(ii) On the two instructional subsets of EgoServe, the system is given the
task-level prior that the benchmarks themselves define. On HoloAssist the
extraction stage receives the declared task type of the recording, its canonical
step sequence, and a short list of known error patterns for that task type. On CaptainCook4D the screening prompt receives the correct step sequence of the recipe being performed, taken from the official task graphs in topological order (at most twenty-five steps). Both are the kind of task knowledge a deployed assistant holds before a session starts, and both follow the setting of these benchmarks, in which error detection is defined relative to a known procedure.
(iii) On the forward-looking tracks of OVO-Bench and ESTP-Bench, where the system must decide \emph{when} to speak rather than answer a question posed afterwards, the candidate user questions of a video are visible to the construction stage; this follows the protocol of these tracks, in which the forward query is known in advance and only its firing time is evaluated. No question is visible during construction on MM-Lifelong, StreamingBench, and backward tracks of OVO-Bench.
 
 
\noindent\textbf{Robustness and Engineering.}
Every LLM call is issued through a wrapper that allows up to three attempts with
exponential backoff, and its output is parsed with a permissive JSON repair pass
before any fallback is taken. The three stages---captioning, segmentation, and
the consolidation of a closed \episode{}---run as an asynchronous pipeline, so
that captioning of later windows overlaps with the consolidation of earlier ones;
within one closed \episode{}, \moment{} extraction and \pattern{} consolidation
are issued concurrently. The resulting memory, its text indices and a resumable
checkpoint are persisted every ten windows, so construction can be interrupted
and continued on a long recording.

\begin{figure*}
    \centering
    \includegraphics[width=1\linewidth]{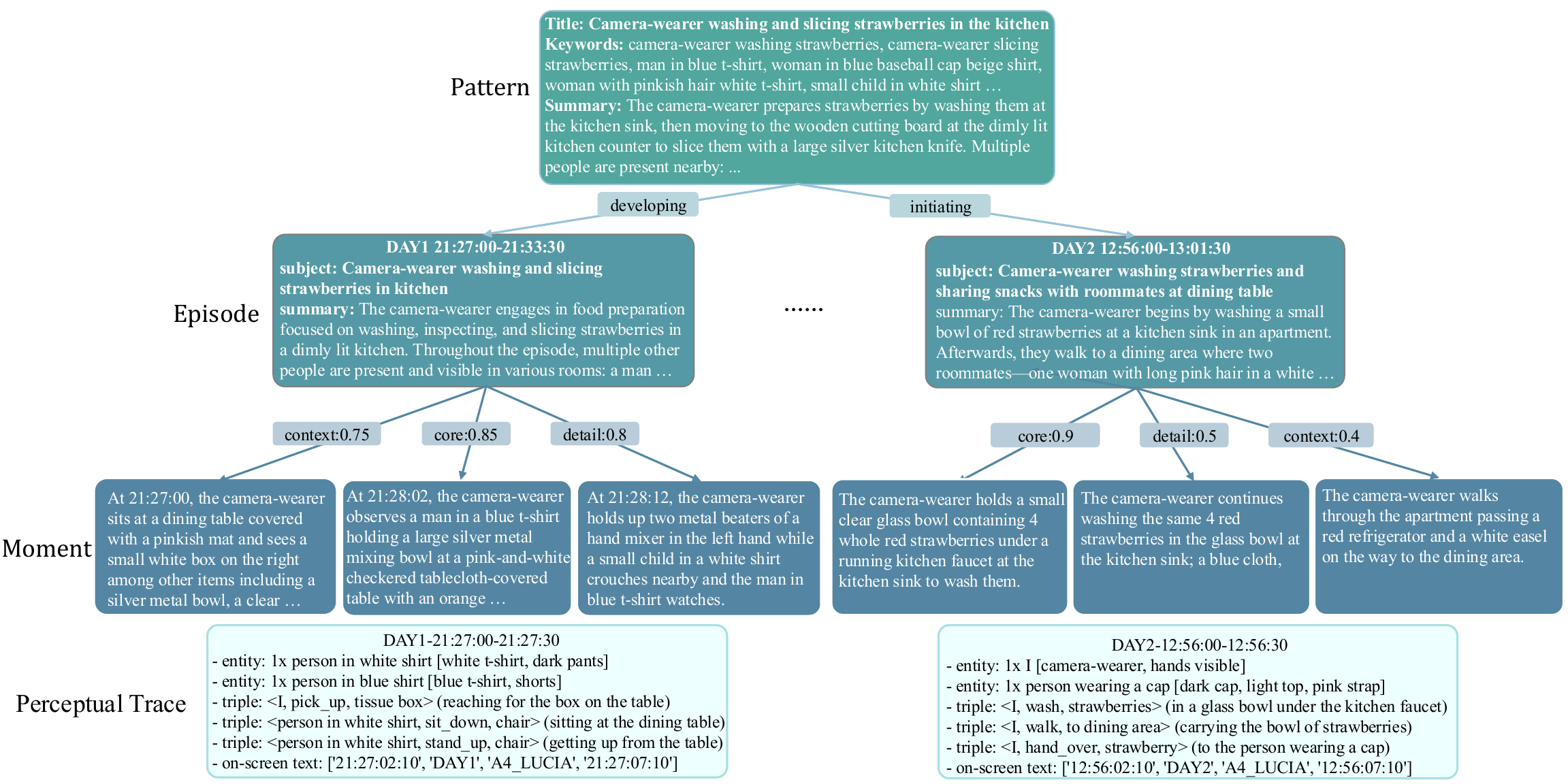}
    \caption{Temporally Stratified Memory Visualization of EgoServe Benchmark}
    \label{fig:case_egoserve}
\end{figure*}

\begin{figure*}
    \centering
    \includegraphics[width=1\linewidth]{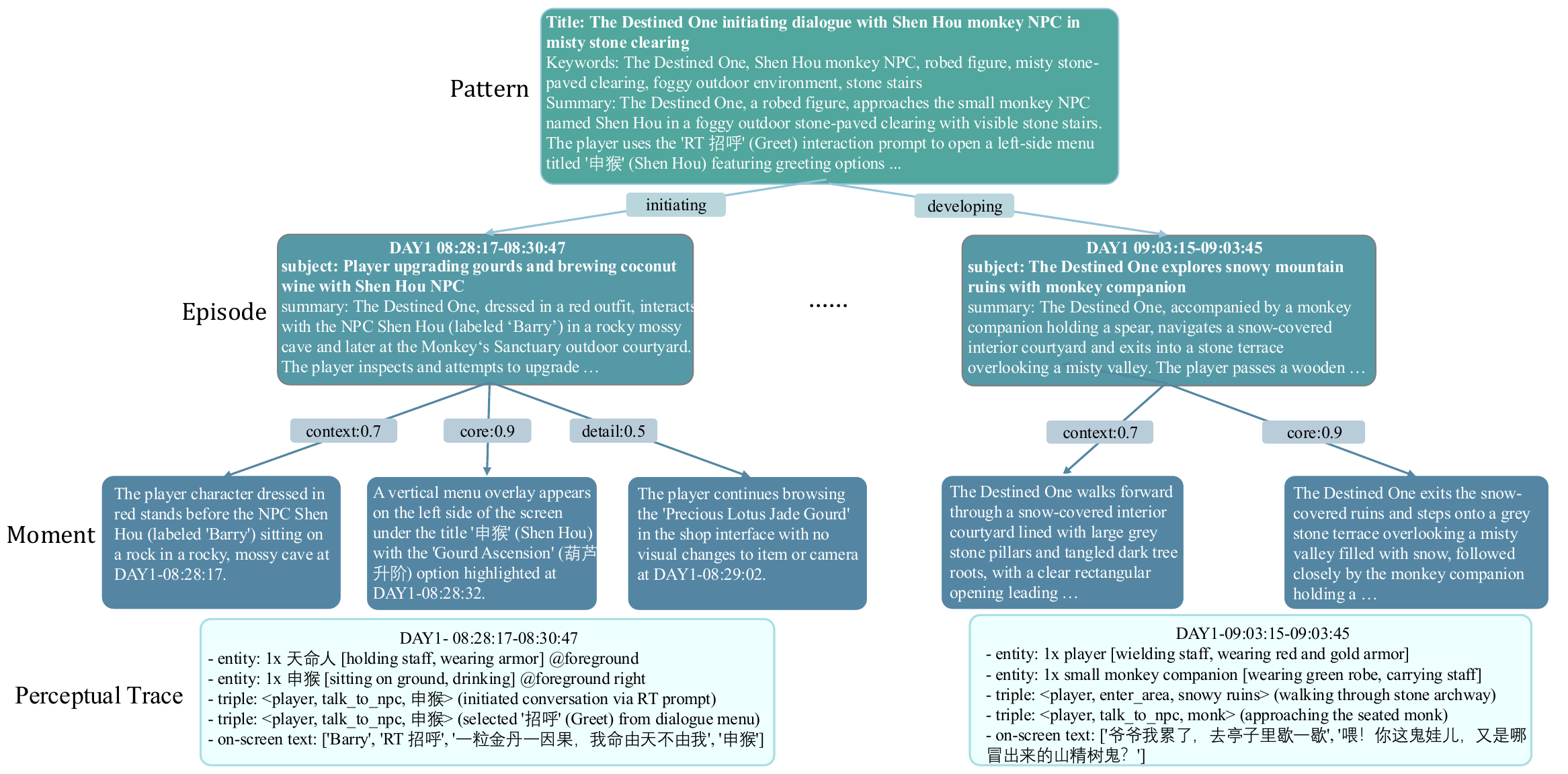}
    \caption{Temporally Stratified Memory Visualization of MMLifelong-Day}
    \label{fig:case_day}
\end{figure*}

\begin{figure*}
    \centering
    \includegraphics[width=1\linewidth]{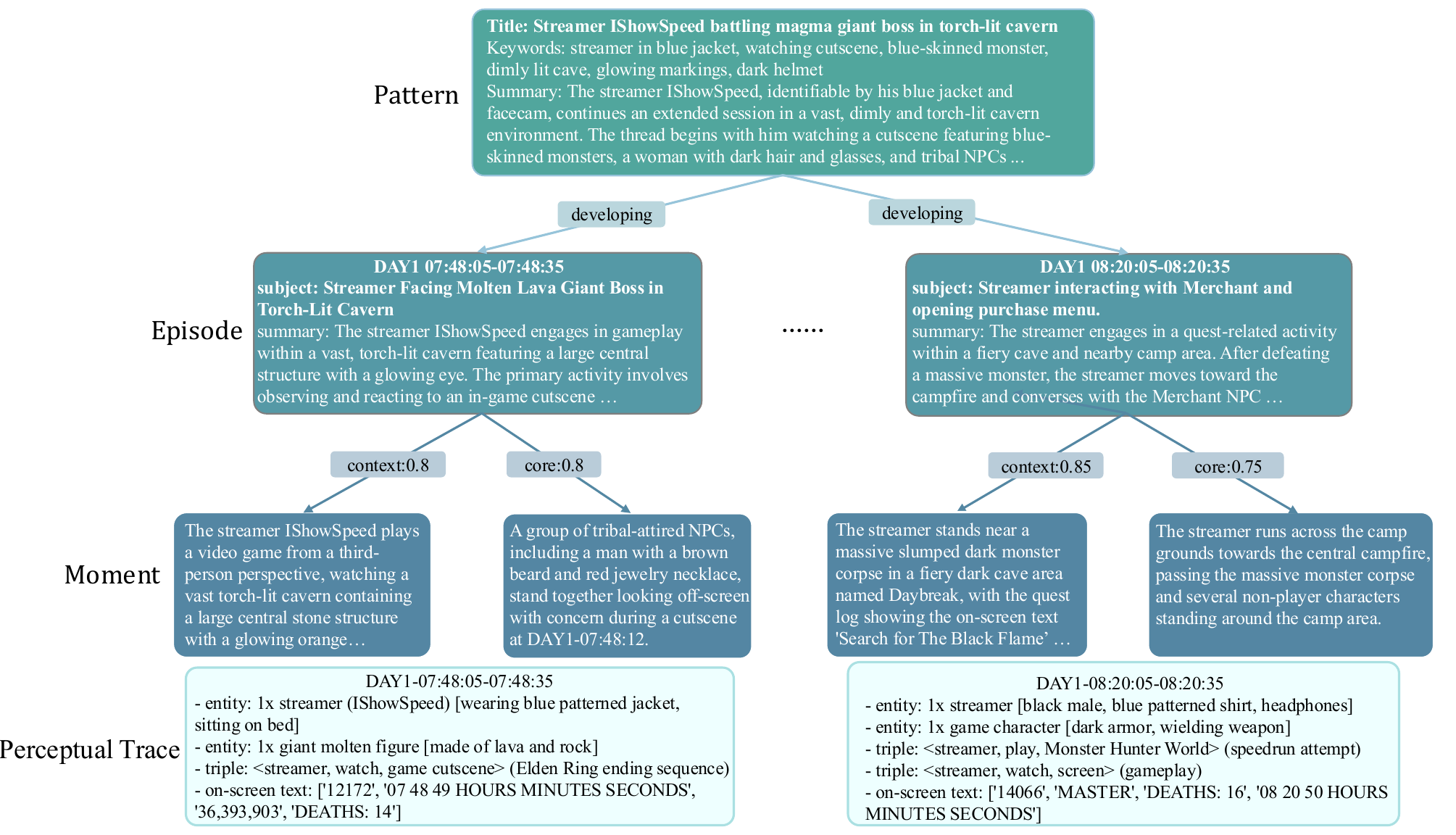}
    \caption{Temporally Stratified Memory Visualization of MMLifelong-Month}
    \label{fig:case_month}
\end{figure*}

\noindent\textbf{Memory Cases Visualization.}
Fig.~\ref{fig:case_egoserve}--\ref{fig:case_month} show one constructed memory per domain, read from the top down: a \pattern{} with its title, keywords and summary, the \episode{}s it groups through a group edge, the \moment{}s each \episode{} expands into with their roles and weights, and the \registry{} entries underneath. A \pattern{} genuinely spans days--in Fig.~\ref{fig:case_egoserve} the same activity is recognized on \texttt{DAY1} and \texttt{DAY2}---and the strata differ in kind rather than only in length, with the \registry{} keeping the literal counts and on-screen text that summaries blur.

\noindent\textbf{Scale of the Constructed Memories.}
Table~\ref{tab:supp_scale_compact} reports the size of the memory built for each
benchmark. Two ratios are informative. First, the \moment{} stratum is an order
of magnitude larger than the \episode{} stratum throughout (from $9.4$
\moment{}s per \episode{} on StreamingBench to $15.2$ on MM-Lifelong), which is
expected of the only stratum that keeps second-level evidence. Second, how much
consolidation the \pattern{} layer performs depends on the horizon rather than on
the amount of video: on the two settings where a single memory spans an entire
recording, $6{,}786$ \episode{}s fold into $2{,}041$ \pattern{}s on MM-Lifelong
and $8{,}002$ into $2{,}531$ on EgoServe (about $3.2$ \episode{}s per
\pattern{}), whereas on the clip-level benchmarks each \pattern{} groups barely
two \episode{}s, since a short clip rarely revisits the same activity. The
\pattern{} stratum therefore grows sub-linearly precisely in the long-horizon
regime it was designed for, which is what keeps the entry point for broad
questions small even after weeks of recording.
 

\begin{table}[t]
    \centering
    \small
    \setlength{\tabcolsep}{4pt}
    \caption{Construction and retrieval hyper-parameters. Values are shared
    across benchmarks unless a row states otherwise; the settings that differ
    follow the length of a recording rather than the domain.}
    \label{tab:supp_hparams}
    \scalebox{0.8}{
    \begin{tabular}{@{}llc@{}}
    \toprule
    Stage & Parameter & Value \\
    \midrule
    \multirow{4}{*}{Perception}
     & Window length (MM-Lifelong, EgoServe) & $30$s \\
     & Window length (other benchmarks) & $10$s \\
     & Frame interval (MM-Lifelong, EgoServe) & $5$s \\
     & Frame interval (other benchmarks) & $2$s \\
    \midrule
    \multirow{5}{*}{Segmentation}
     & Min.\ frames before semantic check & $5$ \\
     & Forced cut at recording gap & $60$s \\
     & \episode{} cap (MM-Lifelong, EgoServe) & $10$min \\
     & \episode{} cap (other benchmarks) & $2$min \\
     & Recent windows shown as anchors & $3$ \\
    \midrule
    \multirow{4}{*}{\moment{}s}
     & Target density ($\le\!5$\,/\,5--15\,/\,$>$15\,min) & $5$\,/\,$10$\,/\,$20$s \\
     & Min.\ \moment{}s per \episode{} & $5$ \\
     & Max.\ \moment{}s per \episode{} & $200$ \\
     & \episode{} summary length & $3$--$4$ sentences \\
    \midrule
    \multirow{4}{*}{\pattern{}s}
     & Candidate batch size & $10$ \\
     & Full summaries kept on re-synthesis & $5$ \\
     & Role\,/\,coherence re-estimation period & $3$ updates \\
     & \pattern{} summary length & $4$--$6$ sentences \\
    \midrule
    \multirow{5}{*}{Retrieval}
     & BM25 $k_1$, $b$ & $1.5$, $0.75$ \\
     & Reciprocal rank fusion $k$ & $60$ \\
     & Traversal fan-out (\pattern{}\,/\,\episode{}\,/\,\moment{}) & $3$\,/\,$8$\,/\,$15$ \\
     & Max.\ rounds $T_{\max}$ (day\,/\,week\,/\,month) & $8$\,/\,$6$\,/\,$12$ \\
     & Max.\ rounds $T_{\max}$ (OVO-Bench, StreamingBench) & $3$\,/\,$2$ \\
    \midrule
    \multirow{2}{*}{LLM calls}
     & Retries with exponential backoff & $3$ \\
     & Decoding & greedy \\
    \bottomrule
    \end{tabular}
    }
\end{table}

\begin{algorithm}[t]
    \caption{Incremental Memory Construction}
    \label{alg:construction}
    \begin{algorithmic}[1]
    \STATE \textbf{Input:} video stream $\{w_1,w_2,\dots\}$; VLM $\Phi$; consolidation LLM $\Psi$
    \STATE \textbf{Init:} $\mathcal{R},\mathcal{F},\mathcal{E},\mathcal{P}\leftarrow\emptyset$; open unit $\tilde{e}\leftarrow\emptyset$
    \FOR{each arriving window $w_t$}
      \STATE $(c_t,r_t)\leftarrow\Phi(w_t)$; \quad $\mathcal{R}\leftarrow\mathcal{R}\cup\{(c_t,r_t)\}$ \COMMENT{queryable at once}
      \IF{recording gap $>60$s \OR day change}
        \STATE $b_t\leftarrow\textsc{cut}$ \COMMENT{deterministic}
      \ELSIF{frames in $\tilde{e}<$ min.\ frames}
        \STATE $b_t\leftarrow\textsc{extend}$ \COMMENT{too early to judge}
      \ELSE
        \STATE $b_t\leftarrow\mathrm{Seg}\big(c_t,\;c_{t-3:t-1}\big)$ 
      \ENDIF
      \IF{$b_t=\textsc{extend}$}
        \STATE absorb $w_t$ into $\tilde{e}$
        \IF{$\mathrm{duration}(\tilde{e})\ge$ cap}
          \STATE $b_t\leftarrow\textsc{cut}$ \COMMENT{length guard}
        \ENDIF
      \ENDIF
      \IF{$b_t=\textsc{cut}$}
        \STATE close $\tilde{e}$ as $e_t$; \quad $\mathcal{E}\leftarrow\mathcal{E}\cup\{e_t\}$
        \STATE \textbf{in parallel:}
        \STATE \quad $\{f_i\},g_{e_t}\leftarrow\Psi(e_t)$; \quad $\mathcal{F}\leftarrow\mathcal{F}\cup\{f_i\}$
        \STATE \quad $p^{*}\leftarrow\mathrm{Match}(e_t,\mathcal{P})$ over concurrent candidate batches
        \STATE \quad $\mathcal{P}\leftarrow\mathrm{Upd}(p^{*},e_t)$ if $p^{*}$ found, else $\mathcal{P}\cup\{\mathrm{New}(e_t)\}$
        \STATE re-open $\tilde{e}\leftarrow\{w_t\}$
      \ENDIF
    \ENDFOR
    \STATE \textbf{Output:} $\mathcal{M}_{\le t}=\{\mathcal{R},\mathcal{F},\mathcal{E},\mathcal{P}\}$, queryable at any $t$
    \end{algorithmic}
    \end{algorithm}

\begin{table}[t]
    \centering
    \small
    \setlength{\tabcolsep}{6pt}
    \caption{Scale of the constructed memories, aggregated per benchmark.
    Each row summarizes the memory structures of all subsets actually used for the reported results. 
    Memories are built per video or per clip except on MM-Lifelong and EgoLife-subset, where one graph spans the whole recording.}
    \label{tab:supp_scale_compact}
    \scalebox{0.9}{
    \begin{tabular}{lcccc}
    \toprule
    Benchmark & Memories & \pattern{}s & \episode{}s & \moment{}s \\
    \midrule
    MM-Lifelong    &     3 & 2{,}041 &  6{,}786 & 102{,}865 \\
    OVO-Bench      & 1{,}640 & 4{,}149 &  8{,}019 &  78{,}497 \\
    StreamingBench &   700 & 3{,}433 & 7{,}128 & 67{,}110 \\
    ESTP-Bench     &   890 & 2{,}675 &  5{,}688 &  73{,}232 \\
    EgoServe    &  283  & 2{,}531 &  8{,}002 &  84{,}866 \\
    \bottomrule
    \end{tabular}
    }
\end{table}

\subsection*{A.2 Retrieval Details on Different Benchmarks}
The four skills of the main paper are abstract interfaces; what a benchmark instantiates depends on how the memory is accessed, which divides the five benchmarks into two groups. On MM-Lifelong, OVO-Bench and StreamingBench, a query opens each round, so the agent selects its own calls under a multi-round budget,
following the skill-based agentic reasoning paradigm. On
ESTP-Bench and EgoServe, no query or a forward request is posed, so retrieval is carried out for the agent by a fixed schema that assembles one evidence block per screening unit, following the schema-based proactive assistance paradigm. 
Table~\ref{tab:supp_toolusage} reports which skills were actually exercised in the runs we report, measured from the inference traces.


\begin{table*}[t]
    \centering
    \small
    \caption{Skills the agent actually invokes in the agentic reasoning stage, in calls per question.
    All figures are measured from the inference traces of the reported runs in the main paper.
    For MM-Lifelong we give both backbones: \texttt{gpt-5-mini}, used throughout the ablations, and \texttt{gpt-5.2}, used for the main-table results. Tools never invoked in a setting are omitted from that column (``--''), so each column reflects what the agent genuinely relies on rather than what the
    framework exposes. Tool names follow the stratum they address
    (\texttt{*\_moment*} for $\mathcal{F}$, \texttt{*\_episode*} for
    $\mathcal{E}$, \texttt{*\_hierarchical} for the
    $\mathcal{P}{\to}\mathcal{E}{\to}\mathcal{F}$ traversal). 
    }
    \label{tab:supp_toolusage}
    \setlength{\tabcolsep}{4pt}
    \begin{tabular}{ll ccc c ccc c cc c}
    \toprule
    & & \multicolumn{7}{c}{\textbf{MM-Lifelong}}
      & \multicolumn{2}{c}{OVO-Bench}
      & StreamingB. \\
    \cmidrule(lr){3-9}\cmidrule(lr){10-11}\cmidrule(lr){12-12}
    & & \multicolumn{3}{c}{\texttt{gpt-5-mini}} & &
        \multicolumn{3}{c}{\texttt{gpt-5.2}} & & & \\
    \cmidrule(lr){3-5}\cmidrule(lr){7-9}
    \multirow{-3}{*}{Skill} & \multirow{-3}{*}{Tool}
      & day & week & month & & day & week & month
      & Real-Time & B.W. & Real-Time \\
    \midrule
    \multirow{6}{*}{$\mathcal{R}$}
      & \texttt{get\_recent\_caption(s)} & --   & --   & --   & & --   & --   & --
        & \textbf{1.05} & --   & \textbf{0.81} \\
      & \texttt{search\_ocr}             & --   & --   & 1.24 & & \textbf{4.38} & 0.41 & 0.96
        & --   & --   & -- \\
      & \texttt{ocr\_in\_range}          & --   & --   & 1.30 & & --   & --   & 0.42
        & --   & --   & -- \\
      & \texttt{search\_events}          & --   & --   & 1.94 & & 0.83 & 0.66 & 1.13
        & --   & --   & -- \\
      & \texttt{count\_events}           & --   & --   & 0.32 & & 0.49 & 0.29 & 0.17
        & --   & --   & -- \\
      & \texttt{search\_entities}        & --   & --   & 0.45 & & 0.03 & 0.43 & 0.11
        & $<$0.01 & \textbf{0.28} & -- \\
    \midrule
    \multirow{3}{*}{$\mathcal{F}$}
      & \texttt{search\_moments}         & \textbf{4.13} & \textbf{2.06} & 2.17 & & 2.06 & 0.89 & 1.83
        & --   & --   & -- \\
      & \texttt{moments\_in\_range}      & 1.15 & 1.46 & 1.56 & & 0.69 & 1.14 & 1.27
        & --   & --   & -- \\
      & \texttt{search\_moment\_only}    & --   & --   & --   & & --   & --   & --
        & --   & 0.11 & -- \\
    \midrule
    \multirow{3}{*}{$\mathcal{E}$}
      & \texttt{search\_episodes}        & 1.49 & 1.37 & \textbf{5.55} & & 1.67 & \textbf{1.91} & \textbf{3.37}
        & --   & --   & -- \\
      & \texttt{get\_recent\_episodes}   & --   & --   & --   & & --   & --   & --
        & 0.01 & 0.07 & -- \\
      & \texttt{search\_episode\_moment} & --   & --   & --   & & --   & --   & --
        & $<$0.01 & 0.23 & -- \\
    \midrule
    $\mathcal{P}{\to}\mathcal{E}{\to}\mathcal{F}$
      & \texttt{search\_hierarchical}    & 2.40 & 1.67 & 0.53 & & 1.09 & 1.24 & 0.33
        & --   & --   & -- \\
    \midrule
    \multicolumn{2}{@{}l}{\textbf{Total calls / question}}
      & \textbf{9.16} & \textbf{6.55} & \textbf{15.07} & & \textbf{11.26} & \textbf{6.96} & \textbf{9.59}
      & \textbf{1.06} & \textbf{0.69} & \textbf{0.81} \\
    \multicolumn{2}{@{}l}{Distinct skills used}
      & 4 & 4 & 9 & & \textbf{8} & \textbf{8} & 9 & 4 & 4 & 1 \\
    \multicolumn{2}{@{}l}{$\mathcal{R}$ share of calls}
      & 0\% & 0\% & 35\% & & \textbf{51\%} & \textbf{26\%} & 29\% & 99\% & 44\% & 100\% \\
    \multicolumn{2}{@{}l}{Accuracy}
      & 18.75 & 19.75 & 17.82 & & \textbf{23.50} & \textbf{22.75} & \textbf{19.98}
      & 76.2 & 69.9 & 77.0 \\
    \bottomrule
    \end{tabular}
\end{table*}

\subsubsection*{Skill-based Agentic Reasoning}

\noindent\textbf{MM-Lifelong.}
The three splits share the same nine calls and the same answer format, and differ
in round budget ($8$/$6$/$12$ for day/week/month), the \pattern{} map
injected as free context before retrieval (capped at $46$ and $100$ entries on
day and week, widened to the full timeline on month), and the content description of the video in the prompt. What the traces reveal is that the same library is exercised very differently. 
As shown in Table~\ref{tab:supp_toolusage}, on day and week, the agent converges on four calls and never issues an OCR, entity, event, or counting one;
on month all nine appear, with the perception stratum taking $35\%$ of the
$15.07$ calls per question. 
The subset also depends on the router rather than the
setting alone: replacing the answer model with \texttt{gpt-5.2} while holding
memory, prompt, and call library fixed widens day and week from four skills to all
eight, and the added calls go almost entirely to $\mathcal{R}$ ($0\%\!\to\!51\%$
of the calls on day, $0\%\!\to\!26\%$ on week), with \texttt{search\_ocr}
becoming the most frequent call on day at $4.38$ per question after never being
issued once by the smaller model. 
Accuracy moves with it ($18.75\!\to\!23.50$ and $19.75\!\to\!22.75$). Skills a weaker router leaves untouched are therefore not dead weight in the library --- they are what a stronger router reaches for first.

\noindent\textbf{OVO-Bench.}
The real-time and backward tracks are answered by a \emph{single} solver: one
tool document, one round prompt, and the same starting context --- the
chronological skeleton of \episode{} summaries with their time spans, covering
every \episode{} that closed before the question moment. The budget is three
rounds with one call each. 
The prompt asks the agent to first decide whether the question concerns the current
moment or the past, and the two profiles that follow are entirely its own. On
\textit{real-time}, the decision resolves to reading the frame-level captions of
the last ten seconds: \texttt{get\_recent\_captions} accounts for $1.05$ of the
$1.06$ calls per question, and the semantic strata are touched fewer than once in
a hundred questions. On \textit{backward}, those captions are never read and the
budget goes to the strata instead --- \texttt{search\_entities} $0.28$,
\texttt{search\_episode\_moment} $0.23$, \texttt{search\_moment\_only} $0.11$,
\texttt{get\_recent\_episodes} $0.07$ per question. 

The \textit{forward} track is different in kind, because the query is known in
advance and only the moment of the response is evaluated. There the agent
receives the recent frame captions as its primary evidence together with the
\moment{}s already retrieved for that query and its own previous answers, and at
each streaming timestamp emits either a response or a decision to keep waiting.

\noindent\textbf{StreamingBench.}
Questions are answered at a given moment of an otherwise unseen stream, so the
instantiation is deliberately perception-first: the \episode{} skeleton up to the
question moment is supplied for free, and the prompt directs perception questions
to call \texttt{get\_recent\_caption} --- the second-level captions of that moment
--- before anything else, leaving the semantic search skills as a fallback for
details that are no longer on screen. Up to three calls may be issued in one
round, which keeps the number of LLM passes low on a benchmark of this size.

The fallback is in practice never taken. On the \textit{real-time} track the
agent issues \texttt{get\_recent\_caption} $0.81$ times per question and nothing
else; across all $2{,}500$ questions the three semantic search skills are invoked
zero times, and roughly one question in five is answered with no call at all,
directly from the skeleton. This is the sharpest instance of a pattern that also
appears on MM-Lifelong: what the agent exercises is decided by what the questions
need, not by what the library offers. The \textit{contextual} track uses the same
instantiation and differs only in that its questions are posed over a longer
preceding span, which is what makes the \episode{} skeleton and the semantic
strata contribute.

\subsubsection*{Schema-based Proactive Assistance}

\noindent\textbf{ESTP-Bench.}
The system must decide \emph{when} and \emph{how} to answer a standing question
rather than answer on demand, so retrieval is carried out by the schema rather
than issued as a call by the agent. At each step of the stream the agent sees the
dense caption of the current step, the interaction history with its own earlier
answers and their timestamps, and --- when the question calls for grounding in
the past --- an evidence block assembled for it from the \moment{} and
\registry{} strata together with the covering \episode{}s.
This block may only ground the content of a response and
never triggers one by itself, which prevents retrieved history from firing an
answer in the absence of present evidence. The two settings reach that block
differently: in the conversational setting it is assembled unconditionally at
every step, whereas in the single-query setting the agent first judges from the
present alone, and the block is assembled only once it asks for grounding, at most
once per question. The budget is therefore deliberately asymmetric --- the present
is always available, the past on request --- and $99.2\%$ of responses are
committed without any retrieval at all, which is what keeps the schema affordable
at streaming rates.

The two settings also differ in what defines a decision step and in how often the
decision is revisited. In the single-query setting, the question stands from the
start of the segment and the decision is revisited at every second-level caption;
in the conversational setting the stream is grouped into short windows that begin
at the moment each turn is asked, carry the dialogue so far, and are revisited
more densely for turns whose answer must track a task in progress. A response is
anchored at a second chosen by the agent and clamped to be causal, and repeated
firings of the same intent are suppressed twice over --- by an instruction not to
restate an answer unless something new appears, and by a hard minimum gap of $8$s
between consecutive responses, tightened to $4$s for questions that ask the system
to comment on each sub-step as it happens.

\noindent\textbf{EgoServe.}
No question is posed at any point, so the agent is driven by a fixed schema
rather than by tool selection. For every screening unit the schema assembles,
from the same strata, the perceived event stream of the unit, the summary of the
covering \episode{}, the \moment{}s inside the unit, the wearer state, and three
kinds of associative candidates retrieved strictly from before the unit ---
entities being re-encountered after hours, earlier requests and object
placements now due, and \pattern{}-level routines. The service taxonomy is given
as a closed set of options with its timing conventions, and the agent returns,
in one pass, any services it decides to fire with an exact trigger second copied
from a perception line.

Its two sources differ in screening unit and in scope, and this is a genuine
configuration difference rather than a prompt variant. The daily-life
recordings (EgoLife) are screened at \episode{} completion --- an adaptive unit,
with a $180$s floor and a $600$s fixed-window fallback --- carry spoken dialogue
and stated intentions, and are judged by three service routers in parallel; since
those routers can independently fire on the same situation, their outputs are
merged by deduplicating per type within $60$s, so that one situation cannot
produce a burst of near-identical prompts. The instructional recordings
(HoloAssist, CaptainCook4D) are screened on fixed short windows of $6$s and $20$s
respectively, without dialogue, and are judged by a single router, so no merge
step is needed; their services concern the task being performed rather than the
wearer's habits, and consequently only the first five of the ten service types
have ground truth there.

\section*{B. More Experimental Results}

\begin{table*}[t]
\centering
\small
\caption{Full comparison on the MM-Lifelong across each subset. 
}
\vspace{-3mm}
\label{tab:supp_mmlifelong}
\scalebox{0.9}
{
\begin{tabular}{lcccc}
\toprule
Method &
Train@Month
&
Val@Month 
& 
Test@Week
& 
Test@Day
\\
\midrule
\textcolor{gray}{Human} & \textcolor{gray}{82.5} & \textcolor{gray}{80.4} & \textcolor{gray}{95.6} & \textcolor{gray}{99.2} \\
\midrule
GPT-5~\cite{singh2025openai} & 10.15 & 14.87 & 15.00 & 15.25 \\
Qwen3-VL-235B-A22B~\cite{bai2025qwen3} & 9.09 & 14.33 & 15.63 & 12.44 \\
Video-XL-2-8B~\cite{qin2025video} & 4.89 & 9.07 & 12.00 & 9.00 \\
Eagle-2.5-8B~\cite{chen2026eagle} & 2.07 & 6.10 & 7.00 & 8.25 \\
Nemotron-v2-12B~\cite{deshmukh2025nvidia} & 7.52 & 9.63 & 11.00 & 7.25 \\
VideoMind-7B~\cite{liu2025videomind} & 5.26 & 8.35 & 11.75 & 7.50 \\
LongVT-7B~\cite{yang2026longvt} & 5.83 & 7.54 & 9.75 & 7.00 \\
DeepVideoDiscovery~\cite{zhang2026deep} & 4.36 & 10.57 & 9.02 & 10.25 \\
ReMA~\cite{chen2026towards} & \underline{17.62} & \underline{18.62} & \underline{18.82} & \underline{16.75} \\
\midrule
\rowcolor{groveblue}
GROVE (ours) & \textbf{18.98} & \textbf{19.98} & \textbf{22.75} & \textbf{23.50} \\
\bottomrule
\end{tabular}
}
\vspace{-3mm}
\end{table*}
\subsection*{B.1 Extended Results of the Main Paper}
\noindent\textbf{Full Comparison on MM-Lifelong.}
Table~\ref{tab:supp_mmlifelong} reports all four splits of MM-Lifelong,
adding the Train@Month split that the main paper omits for space.
\method{} attains the best score on every split, and its margin over ReMA
is preserved on the month-scale training videos ($18.98$ vs.\ $17.62$), indicating that the gains come from how the memory is organized and retrieved rather than from any single evaluation split. 

\noindent\textbf{Memory Structure on the Month Split.}
As illustrated in Table~\ref{tab:supple_abl_structure}, the scale dependence observed on day and week continues to grow with the
horizon. 
Removing \pattern{} costs the most of any single stratum on month
(17.82$\to$14.85), a larger drop than on week, since a question spanning a 22-day recording is usually answered by a recurring activity rather than by one occurrence. 
Flattening the hierarchy is again the worst configuration
(13.56) and its gap to \method{} is the widest of the three splits, whereas removing \episode{} costs little here (17.74) in contrast to day (14.57): month-scale questions rarely hinge on localizing one segment. The \registry{} is likewise less critical at this scale (18.62), as the third-person travelling recordings contain fewer of the exact-count and on-screen-text cues it preserves.

\noindent\textbf{Retrieval Skills on the Month Split.}
\emph{Moment Recall} remains the skill the agent cannot do without
(17.82$\to$15.16), consistent with the week split. 
Two entry points behave differently from the shorter horizons: removing \emph{Episode Replay} slightly improves accuracy (18.22), because searching several hundred \episode{} summaries often returns plausible but wrong segments and consumes rounds that the chapter map would have spent better; and removing \emph{Pattern Traversal} is nearly neutral (17.90), since month-scale counting and ordering questions are already answered by enumerating occurrences rather than by a top-down traversal. Disabling retrieval altogether still collapses accuracy (11.16), confirming that the gains come from the retrieved evidence and not from the reasoning backbone alone.

\noindent\textbf{Memory Structure across Service Types.}
Table~\ref{tab:supple_egoserve1} breaks the EgoServe ablation into the ten service
sub-types, which makes visible what the aggregate macro-F1 hides: each stratum
supports a different band of the service spectrum. 
The two informative cases are
\pattern{} and \moment{}, which have opposite profiles. Removing \pattern{} leaves
the instant and short-term families untouched (NSG $21.8\!\rightarrow\!21.6$, ER
$23.8\!\rightarrow\!24.1$) while precisely those services that must recognize a
\emph{recurring} activity collapse --- memory recall $5.9\!\rightarrow\!2.7$, task
reminder $5.9\!\rightarrow\!4.5$, routine optimization $7.1\!\rightarrow\!4.2$ ---
and it is the only ablation whose damage falls mainly on the far end of the
spectrum ($-1.74$ on average there against $-0.54$ at the near end, every other
stratum being the other way round). A reminder that an activity is due can only be
issued if that activity has been recognized across days, whereas a safety alert
needs nothing beyond the current window. \moment{} is the mirror image and the
costliest stratum at the near end ($-2.38$ on average), hurting tool use
($20.2\!\rightarrow\!16.7$) and next-step guidance ($21.8\!\rightarrow\!17.1$)
most; it also carries habit coaching ($8.6\!\rightarrow\!4.3$), which survives the
\pattern{} ablation ($8.6\!\rightarrow\!8.2$) but not this one, since a
second-level record is what lets a service be both triggered and timed. 
The remaining rows behave as expected: the \registry{} concentrates its damage on the
instant family (SA $19.7\!\rightarrow\!14.9$) and flattening the memory is worst on
the types that need organization (SA $13.5$, ML $4.4$). 

\begin{table}[t]
\centering
\caption{Ablation of memory structure on the three MM-Lifelong splits. (day/week/month)}
\label{tab:supple_abl_structure}
\vspace{-3mm}
\scalebox{0.9}{
\begin{tabular}{l ccc}
\toprule
Config & day & week & month \\
\midrule
\rowcolor{groveblue}
\textbf{GROVE} & \underline{18.75} & \underline{19.75} & \underline{17.82} \\
\midrule
w/o Perceptual Trace & 15.50 & \textbf{20.50} & \textbf{18.62} \\
w/o Moment & 17.09 & 18.25 & 16.53 \\
w/o Episode & 14.57 & 17.25 & 17.74 \\
w/o Pattern & \textbf{19.90} & 13.00 & 14.85 \\
Flat (w/o Hier.) & 14.75 & 15.50 & 13.56 \\
\bottomrule
\end{tabular}
}
\vspace{-3mm}
\end{table}
\begin{table}[t]
\centering
\caption{Ablation of retrieval skills on the three MM-Lifelong splits. (day/week/month)}
\vspace{-3mm}
\label{tab:supple_abl_primitive}
\scalebox{0.9}{
\begin{tabular}{l ccc}
\toprule
Config & day & week & month \\
\midrule
\rowcolor{groveblue}
\textbf{GROVE} & \underline{18.75} & \underline{19.75} & 17.82 \\
\midrule
w/o Perception Lookup & 16.25 & \textbf{20.25} & 16.45 \\
w/o Moment Recall & \textbf{19.50} & 15.75 & 15.16 \\
w/o Episode Replay & 18.34 & 19.60 & \textbf{18.22} \\
w/o Pattern Traversal & 18.50 & 15.58 & \underline{17.90} \\
No retrieval & 7.25 & 12.75 & 11.16 \\
\bottomrule
\end{tabular}
}
\vspace{-3mm}
\end{table}
\begin{table*}[t]
\centering
\caption{Memory structure ablation results of each sub-type services on EgoServe benchmark. }
\vspace{-3mm}
\label{tab:supple_egoserve1}
\scalebox{0.9}{
\begin{tabular}{l|cc|ccc|cc|ccc|c}
\toprule
\multirow{2}{*}{Model} & \multicolumn{2}{c|}{Instant} & \multicolumn{3}{c|}{Short-term} & \multicolumn{2}{c|}{Episodic} & \multicolumn{3}{c|}{Long-term} & \multirow{2}{*}{Overall} \\
 & SA & TU & NSG & ER & RR & MR & TR & HC & ML & RO & \\
\midrule
\rowcolor{groveblue}
GROVE (Ours) & \textbf{19.7} & \underline{20.2} & \textbf{21.8} & 23.8 & 6.8 & \underline{5.9} & 5.9 & \textbf{8.6} & 6.5 & 7.1 & \textbf{12.62} \\
\midrule
w/o Perceptual Trace &14.9 &18.8 &20.0 &22.6 &\textbf{11.2} &3.7 &\textbf{9.8} &4.4 &\underline{8.2} &9.2 &12.27 \\
w/o Moment &\underline{17.6} &16.7 &17.1 &22.7 &6.3 &4.5 &\underline{6.2} &4.3 &\textbf{9.4} & \textbf{12.7} &11.77 \\
w/o Episode &16.9 & \textbf{20.6} &21.2 &22.8 &4.8 &\textbf{8.2} &7.3 &5.6 &5.4 &\underline{11.8} & \underline{12.47} \\
w/o Pattern &16.8 &19.4 &\underline{21.6} &\underline{24.1} &7.7 &2.7 &4.5 &\underline{8.2} &5.7 &4.2 &11.49 \\
Flat &13.5 &21.5 &18.0 &\textbf{24.8} &\underline{6.9} &4.0 &7.3 &5.6 &4.4 &9.3 &11.53 \\
\bottomrule
\end{tabular}}
\vspace{-3mm}
\end{table*}
\begin{table*}[t]
\centering
\caption{Retrieval skill ablation results of each sub-type services on the EgoServe benchmark. }
\vspace{-3mm}
\label{tab:supple_egoserve2}
\scalebox{0.9}{
\begin{tabular}{l|cc|ccc|cc|ccc|c}
\toprule
\multirow{2}{*}{Model} & \multicolumn{2}{c|}{Instant} & \multicolumn{3}{c|}{Short-term} & \multicolumn{2}{c|}{Episodic} & \multicolumn{3}{c|}{Long-term} & \multirow{2}{*}{Overall} \\
 & SA & TU & NSG & ER & RR & MR & TR & HC & ML & RO & \\
\midrule
\rowcolor{groveblue}
GROVE (Ours) & \textbf{19.7} & \textbf{20.2} & \textbf{21.8} & 23.8 & 6.8 & \textbf{5.9} & 5.9 & \textbf{8.6} & 6.5 & 7.1 & \textbf{12.62} \\
\midrule
w/o Perception Lookup&14.9 &18.8 &20.0 &22.6 &\textbf{11.2} &3.7 &\textbf{9.8} &4.4 &\underline{8.2} &\textbf{9.2} &\underline{12.27} \\
w/o Moment Recall &13.9 &18.5 &16.5 &\textbf{24.9} &\underline{9.4} &\underline{4.9} &\underline{6.9} &2.8 &5.7 &6.8 &11.05 \\
w/o Episode Replay &14.1 &18.0 &21.2 &23.6 &\underline{8.5} &1.1 &5.8 &\underline{8.5} &\textbf{8.7} &4.0 &11.36 \\
w/o Pattern Traversal & \underline{16.8} & \underline{19.4} & \underline{21.6} &24.1 &7.7 &2.7 &4.5 &8.2 &5.7 &4.2 &11.49 \\
No retrieval &15.1 &17.4 &16.5 &\underline{24.2} &6.6 &1.6 &4.2 &4.2 &6.5 &\underline{7.6} &10.39 \\
\bottomrule
\end{tabular}}
\vspace{-3mm}
\end{table*}

\noindent\textbf{Retrieval Skills across Service Types.}
Table~\ref{tab:supple_egoserve2} shows that an entry point is only as useful as
the band it serves. \emph{Moment Recall} is the most consequential skill overall
($12.62\!\rightarrow\!11.05$); its removal is felt on safety alerts
($19.7\!\rightarrow\!13.9$), next-step guidance ($21.8\!\rightarrow\!16.5$) and
habit coaching ($8.6\!\rightarrow\!2.8$) --- the three types whose decision needs
a specific second rather than a general situation. The services that reach further
back instead depend on the two coarse-grained skills: without \emph{Episode
Replay} memory recall almost vanishes ($5.9\!\rightarrow\!1.1$) and routine
optimization drops ($7.1\!\rightarrow\!4.0$), because a recalled fact is only
actionable when the activity that surrounds it can be read; without \emph{Pattern
Traversal} memory recall ($2.7$) and routine optimization ($4.2$) fall to exactly
the level seen when the \pattern{} stratum itself is deleted, since on this
benchmark the two interventions coincide. With no retrieval at all the pattern is
unmistakable: safety alerts remain usable ($15.1$) because they are decided from
the present scene, while memory recall ($5.9\!\rightarrow\!1.6$), task reminder
($5.9\!\rightarrow\!4.2$) and habit coaching ($8.6\!\rightarrow\!4.2$) lose most
of their accuracy. Proactive assistance that merely reacts to what is visible
needs no memory; assistance that refers to what happened earlier is memory-bound.


\begin{table}[t]
\centering
\caption{Ablation results of retrieval width on the MMLifelong. }
\label{tab:supp_topk}
\vspace{-3mm}
\begin{tabular}{l ccc}
\toprule
Config & day & week & month \\
\midrule
 top-$k$=3 & 13.25 & 15.50 & 14.84 \\
 top-$k$=10 & \underline{16.50} & 16.08 & 16.16 \\
 \rowcolor{groveblue}
 top-$k$=20 (Ours) & \textbf{18.75} & \textbf{19.75} & \underline{17.82} \\
 top-$k$=30 & 16.33 & \underline{19.50} & \textbf{17.98} \\
\bottomrule
\end{tabular}
\end{table}
\begin{table}[t]
\centering
\caption{Ablation on how retrieval skills are selected. 
\emph{Only BM25} keeps the full skill set but disables the semantic branch of the cross-stratum traversal, so ranking is lexical. \emph{Only Moments / Episodes / Hierarchical} force every call to resolve to that single skill. \emph{Random-1} and \emph{Random-3} replace each chosen skill with one or three sampled uniformly at random. \emph{Fixed All Tools} executes every skill on every round. 
Its round budget is capped at $3$, the setting whose call volume comes closest to GROVE's.
}
\label{tab:supp_toolrouting}
\vspace{-3mm}
\begin{tabular}{l ccc}
\toprule
Config & day & week & month \\
\midrule
 Only BM25 & 17.75 & 15.00 & 16.69 \\
 \hdashline[2pt/2pt]
 Only Moments & 12.75 & 17.00 & \underline{16.77} \\
 Only Episodes & 12.00 & 14.50 & 14.53 \\
 Only Hierarchical & 13.00 & 15.25 & \underline{16.77} \\
 \hdashline[2pt/2pt]
 Random-1 & 16.00 & 15.75 & 16.53 \\
 Random-3 & \textbf{20.75} & 18.25 & 15.97 \\
 Fixed All Tools & 18.00 & \underline{18.75} & 16.05 \\
 \hdashline[2pt/2pt]
 \rowcolor{groveblue}
 GROVE (Ours) & \underline{18.75} & \textbf{19.75} & \textbf{17.82} \\
\bottomrule
\end{tabular}
\end{table}
\begin{table*}[t]
\centering
\caption{Ablation of Adaptive Segmentation of each sub-type service on EgoServe.}
\vspace{-3mm}
\label{tab:egoserve_adaptive_seg}
\scalebox{0.9}{
\begin{tabular}{l|cc|ccc|cc|ccc|c}
\toprule
\multirow{2}{*}{Model} & \multicolumn{2}{c|}{Instant} & \multicolumn{3}{c|}{Short-term} & \multicolumn{2}{c|}{Episodic} & \multicolumn{3}{c|}{Long-term} & \multirow{2}{*}{Overall} \\
 & SA & TU & NSG & ER & RR & MR & TR & HC & ML & RO & \\
\midrule
\rowcolor{groveblue}
GROVE (Ours) & \textbf{19.7} & \textbf{20.2} & \textbf{21.8} & \underline{23.8} & \underline{6.8} & \textbf{5.9} & \textbf{5.9} & \textbf{8.6} & \underline{6.5} & \underline{7.1} & \textbf{12.62} \\
\midrule
w/o Adaptive Segmentation &\underline{14.1} & \underline{19.4} & \textbf{21.8} & \textbf{23.9} & \textbf{9.3} & \underline{4.5} & \underline{5.5} & \underline{4.2} & \textbf{8.3} & \textbf{10.7} & \underline{12.17} \\
\bottomrule
\end{tabular}}
\vspace{-3mm}
\end{table*}

\subsection{B.2 Additional Experimental Results}

\noindent\textbf{Effect of the Retrieval Width.}
Table~\ref{tab:supp_topk} varies the number of items each skill returns per call, top-$k$, while keeping the memory, the skills, and the round budget unchanged. A narrow setting starves the agent: at $k\!=\!3$ every split loses
three to five points, because the evidence that answers the question is often
not among the first few ranked items. Accuracy peaks at $k\!=\!20$ on the day
and week splits and then degrades on day ($18.75\rightarrow16.33$), where the
extra items are mostly redundant and dilute the context the agent has to read.
The month split is the exception, improving marginally up to $k\!=\!30$
($17.82\rightarrow17.98$): questions spanning weeks require evidence gathered from more places at once, so a wider return is still useful.

\noindent\textbf{How Skills Are Selected.}
Table~\ref{tab:supp_toolrouting} keeps the memory, the prompt, and the round budget
fixed and changes only how calls are chosen; the agent still decides how many to
issue, so the substituted rows spend a comparable budget ($9.2$ calls per
question for GROVE against $9.8$--$10.4$). 
\emph{Only BM25} keeps the traversal
intact and disables just its semantic branch: day ($17.75$) is unaffected because the questions quote words that appear verbatim in the memory, but week drops to $15.00$ once a
question and the recorded evidence are worded differently --- semantics matters
only once the memory is large enough for lexical overlap to become ambiguous.
The \emph{Only} rows force every call onto one stratum and are the most damaging
intervention, costing close to six points on day; more tellingly, their ranking
inverts across horizons (\emph{Only Moments} is the weakest on day at $12.75$
yet the strongest on week at $17.00$), so no fixed choice of entry point
transfers, and even the best of these single-stratum baselines ($13.00$) trails
random alternation among all of them ($16.00$). The last block removes the choice without removing the
retrieval: \emph{Random-3} wins on day ($20.75$) but only because concatenating
three returns inflates the evidence block by ${\sim}40\%$ in prompt tokens, and
that advantage vanishes as the horizon grows ($15.97$ on month); \emph{Fixed All
Tools} retrieves everything every round at $1.7\times$ GROVE's call volume and
still loses on all three splits. Selecting where to look is therefore what the
longer horizons reward, and the margin over \emph{Fixed All Tools} widens from
$+0.75$ on day to $+1.77$ on month.

\noindent\textbf{Effect of Adaptive Segmentation.}
Table~\ref{tab:egoserve_adaptive_seg} isolates the segmenter on EgoServe. We keep
the segment-level evidence block but redraw its boundaries: instead of the content-driven \episode{} boundaries, the same span is cut into the \emph{same number} of equal-length segments, each filled with the captions it covers, with the count matched per recording (and per day on EgoLife, where the cuts are drawn inside the spans that contain video so that sleep and away periods do not consume segments). The screening windows, the segment count and the evidence volume are therefore unchanged, and the only variable is where the boundaries fall. Removing adaptive segmentation costs $12.62\!\rightarrow\!12.17$ overall, and the loss concentrates on the two services whose decision must be anchored to a particular second: safety alerts drop $19.7\!\rightarrow\!14.1$ and habit coaching $8.6\!\rightarrow\!4.2$, in both cases because the number of matched predictions falls outright ($49\!\rightarrow\!35$ and $6\!\rightarrow\!3$) at an unchanged prediction volume --- the responses are still issued, but they no longer land inside the tolerance window. A fixed-length unit no longer begins and ends where the activity does, so the summary grounding the decision mixes two activities and the trigger second drifts from the event it should mark.


\begin{figure*}
    \centering
    \includegraphics[width=1\linewidth]{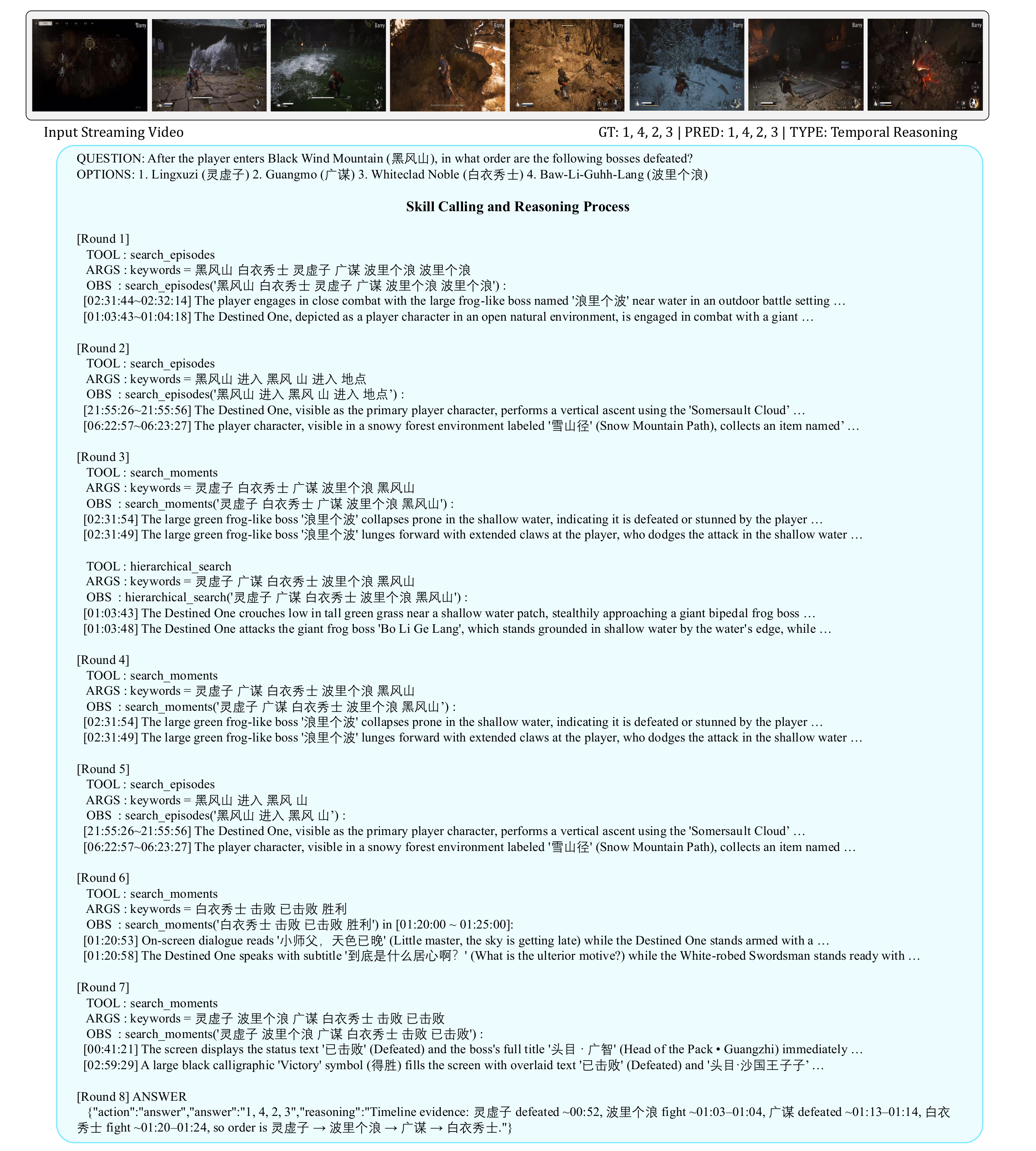}
    \caption{Reasoning and Skill Scale Calling Example on the MMLifelong (day). }
    \label{fig:case_trace_day}
\end{figure*}

\section*{C. Case Study}
Fig.~\ref{fig:case_trace_day} shows the full skill-calling and reasoning trace of GROVE on a temporal-ordering question on the MMLifelong day subset.
The four bosses are defeated across a $32$-minute span of a $24.5$-hour
recording, so no single call can cover more than one of them. GROVE first
locates the region with \texttt{search\_episodes} (rounds~1--2), then combines
\texttt{search\_moments} with the $\mathcal{P}{\to}\mathcal{E}{\to}\mathcal{F}$
traversal to pull the individual defeat records (round~3), and finally narrows
to the time ranges those rounds exposed --- e.g.\ \texttt{search\_moments} in
$[01{:}20{:}00,01{:}25{:}00]$ at round~6 --- until every boss carries a
timestamp. The answer is read off the reconstructed timeline rather than
recalled, giving the correct order \texttt{1,\,4,\,2,\,3}.

\bibliography{references}


\end{document}